%% file: iclr2027_arxiv.tex
\documentclass{article} 
\usepackage{iclr2027_conference,times}

\input{math_commands.tex}

\usepackage{hyperref}
\usepackage{url}
\usepackage{booktabs}
\usepackage{xspace}
\usepackage{multirow}
\usepackage{tikz}
\usepackage{graphicx}
\usepackage{subcaption}
\usepackage{enumitem}
\usepackage{wrapfig} 

\usepackage{listings}
\lstdefinestyle{pythonpseudo}{
    language=Python,
    alsoletter={.}, 
    basicstyle=\ttfamily\scriptsize,
    keywordstyle=\bfseries,
    commentstyle=\itshape,
    stringstyle=\ttfamily,
    numberstyle=\tiny,
    numbers=left,
    numbersep=6pt,
    frame=single,
    framerule=0.4pt,
    breaklines=true,
    keepspaces=true,
    showstringspaces=false,
    xleftmargin=22pt,
    xrightmargin=2pt,
    aboveskip=6pt,
    belowskip=6pt,
}

\title{MOMO: Human Motion from Moving Sounds}

\author{Deli Cai$^{1}$ \;
    Shuyang Xu$^{2}$ \; 
    Zhiyang Dou$^{3}$ \;
    Zitong Lan$^{4}$ \; \\
    \textbf{Mingmin Zhao$^{4}$ \;
    Changxing Ding$^{1}$\thanks{Corresponding Author} \;
    Taku Komura$^{2}$\footnotemark[1]} 
    \\
    \textsuperscript{1}South China University of Technology \; 
    \textsuperscript{2}The University of Hong Kong \; 
    \\
    \textsuperscript{3}MIT \; 
    \textsuperscript{4}University of Pennsylvania}

\newcommand{\name}{MOMO\xspace}
\newcommand{\dbname}{DySAM\xspace}
\newcommand{\dbnamessl}{DySAL\xspace}

\newcommand{\remove}[1]{}

\newcommand{\circled}[1]{
    \tikz[baseline=(char.base)]{
        \node[shape=circle, draw, inner sep=0pt, outer sep=0pt, minimum size=1pt] (char) {#1};
    }
}
\newcommand{\et}[2]{${#1}^{\pm{#2}}$} 
\newcommand{\etb}[2]{$\mathbf{{#1}}^{\pm{#2}}$} 

\usepackage{amssymb} 
\newcommand{\Checkmark}{\checkmark} 

\iclrfinalcopy 
\begin{document}

\maketitle

\begin{figure}[!ht]
    \vspace{-9mm}
    \centering
       \includegraphics[width=\linewidth]{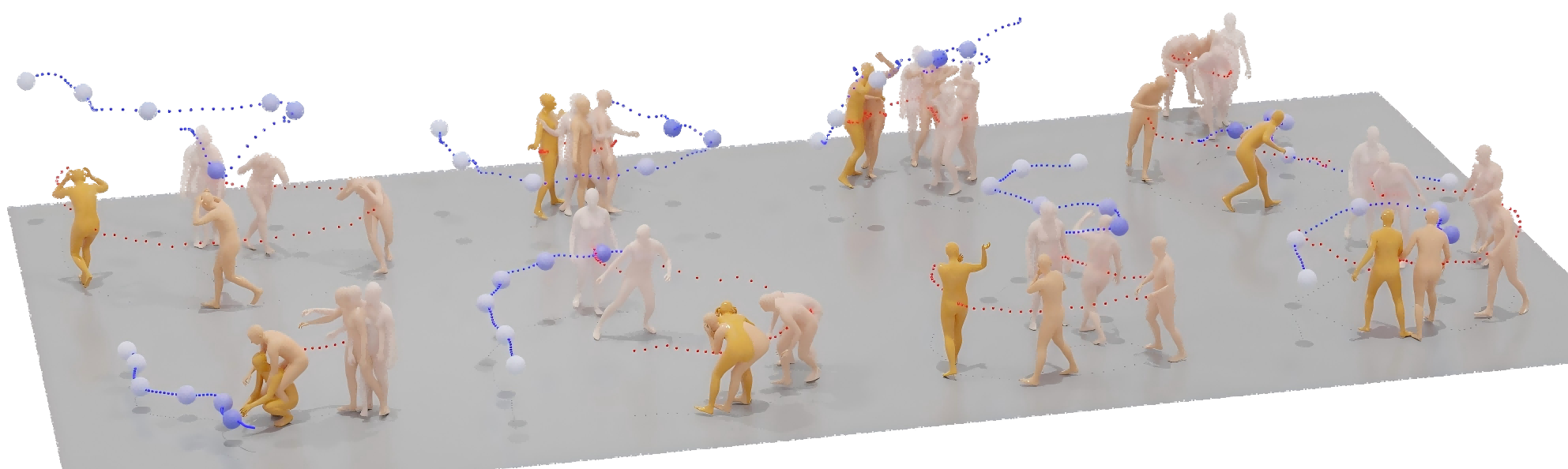}
       \caption{This paper proposes a novel task for synthesizing 3D human motions in response to \underline{\textit{dynamic}} \underline{\textit{spatial}} \textit{audio}. To support this new task, we collect \dbname, the first dataset comprising synchronized dynamic spatial audio, corresponding sound source trajectories, and high-quality 3D human motion. In this figure, the model generates diverse and realistic human reactions to various moving acoustic events.}
    \vspace{0mm}
    \label{fig:teaser}
\end{figure}

\input{sec/0_abstract.tex}

\input{sec/1_intro.tex}

\input{sec/2_related.tex}

\input{sec/3_dataset.tex}

\input{sec/4_method.tex}

\input{sec/5_experiment.tex}

\input{sec/6_conclusion.tex}

\subsection*{AI use statement}

In this work, we used generative AI tools for assisting with code implementation, editing, completion, and debugging. 
We have not used generative AI tools for generating synthetic datasets, developing theoretical models or conceptual frameworks, formulating mathematical claims, providing critical ingredients for or assisting in writing proofs, proposing or refining research hypotheses, designing or providing feedback on research methodology or experiments, assisting with translation, cleaning or reformatting datasets, supporting qualitative or thematic data analysis, or interpreting experimental results, and the remaining required disclosure tasks are not applicable to this work. 
Additionally, we used generative AI tools for drafting and editing parts of the manuscript, improving grammar, clarity, organization, and readability, and assisting with general software code creation or editing. 
We have reviewed all AI-assisted work. 
All AI-assisted manuscript text was carefully reviewed and revised by the authors to ensure factual and technical accuracy and to preserve the authors' intended meaning. 
All AI-assisted code was manually inspected, executed, tested, and validated by the authors before being used in the experiments. 
We take responsibility for the final content of this work, including text, claims or artifacts produced with the aid of generative AI.

\subsection*{Reproducibility Statement}

To support reproducibility, we provide detailed descriptions of the dataset construction, model architecture, implementation details, training procedure, and evaluation protocols in the main paper and appendix. Implementation details are included in the supplementary material. Upon acceptance, we will publicly release the datasets, source code, pretrained models, and data-processing scripts to facilitate reproduction and future research.

\bibliography{iclr2027_conference}
\bibliographystyle{iclr2027_conference}

\input{sec/7_appendix.tex}

\end{document}

%% file: math_commands.tex
\usepackage{amsmath,amsfonts,bm}

\def\eqref#1{equation~\ref{#1}}

\def\1{\bm{1}}

\DeclareMathAlphabet{\mathsfit}{\encodingdefault}{\sfdefault}{m}{sl}
\SetMathAlphabet{\mathsfit}{bold}{\encodingdefault}{\sfdefault}{bx}{n}



%% file: sec/0_abstract.tex
\begin{abstract}

Human responses to sound depend not only on its semantic content, but also on the evolving spatial relationship between source and listener, conveyed through {spatial audio} and its {temporal dynamics}. Modeling human reactions to such dynamic spatial audio is important for character animation, human behavior analysis, and robotics. As of yet, existing audio-driven motion generation methods primarily focus on non-spatial or static acoustic conditions, leaving this important problem largely underexplored. To bridge this gap, we formulate the novel task of dynamic spatial-audio-driven human motion generation and introduce two motion-captured datasets. We introduce the \textbf{Dy}namic \textbf{S}patial \textbf{A}udio and \textbf{M}otion dataset (\dbname) as the first dataset and benchmark for this task, providing synchronized binaural audio, 3D human motions, and moving sound source trajectories. Given the importance of dynamic spatial cues of sound sources, we further provide the \textbf{Dy}namic \textbf{S}patial \textbf{A}udio \textbf{L}ocalization dataset (\dbnamessl), an auxiliary binaural localization dataset that pairs diverse acoustic events with head poses and moving source trajectories to support spatial audio perception.
For benchmarking, we propose \name{} for human \textbf{Mo}tion generation from \textbf{Mo}ving Sounds, a diffusion-based framework guided by three coupled principles: localization, understanding, and generation.
\name performs localization by estimating head-centric spatial direction cues of the moving source directly from raw binaural audio, avoiding reliance on ground-truth source coordinates. It performs understanding by extracting frame-wise audio-content representations, and performs generation by conditioning motion diffusion on both semantic features and dynamic spatial cues. Moreover, \name incorporates frame-wise motion correction during denoising, enabling temporally adaptive motion synthesis under evolving audio-spatial cues.
Experiments show that \name outperforms existing baselines in motion fidelity, audio-motion alignment, and physical plausibility, demonstrating the potential of dynamic spatial audio for embodied motion generation.
Project page: \url{https://cdlchoi.github.io/projects/MOMO}

\end{abstract}

%% file: sec/1_intro.tex
\newpage
\section{Introduction}

\begin{wrapfigure}{r}{0.3\textwidth}
    \centering
    \vspace{-3mm}
    \includegraphics[width=\linewidth]{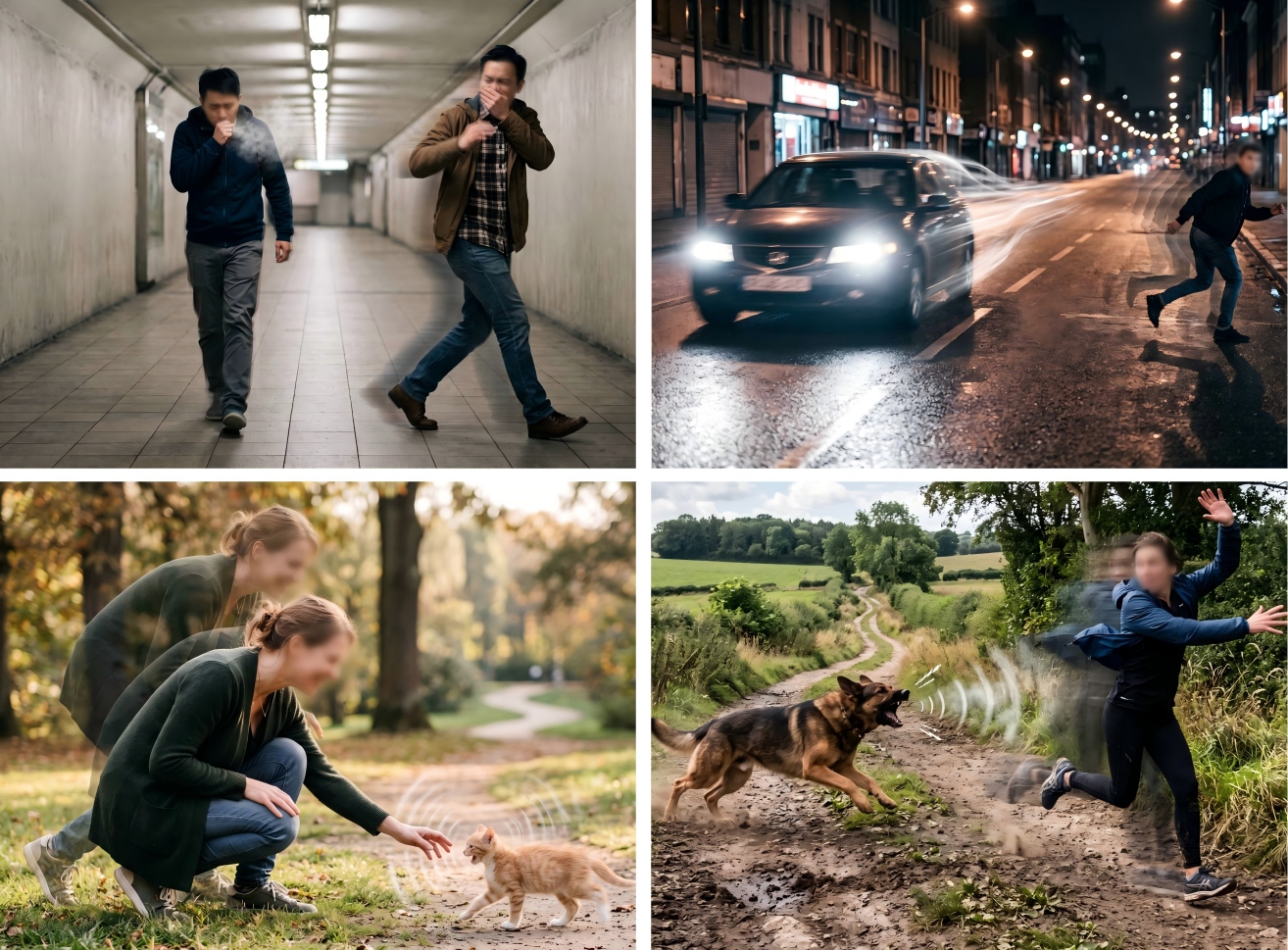}
    \vspace{-6mm}
    \caption{Diverse Reactions to Moving Sounds.}
    \vspace{-4mm}
    \label{fig:real_world}
\end{wrapfigure}
Human responses to sound arise from integrating semantic meaning with \textit{spatial audio} cues and the \textit{temporal dynamics of sound sources}. A barking dog approaching from behind may elicit avoidance, whereas a speeding car approaching may prompt reorientation. Crucially, the same sound can evoke different actions depending on its source's location and motion relative to the listener. Modeling this perception–action coupling supports context-aware virtual humans and embodied agents in character animation, virtual reality, human–computer interaction, and embodied intelligence.

Existing human motion datasets and methods have mainly focused on conditions such as action labels \citep{guo2020action2motion}, text descriptions \citep{guo2022generating}, music \citep{zhuang2022music2dance,li2021ai,tseng2023edge,li2023finedance,li2024lodge,luo2024popdg}, and speech \citep{ao2022rhythmic,ao2023gesturediffuclip,alexanderson2023listen}. Recently, to the best of our knowledge, MOSPA \citep{xu2025mospa} introduced the first spatial-audio-driven human motion dataset, SAM, along with a corresponding motion generation method, showing that spatial audio cues are crucial for generating realistic human motions. However, the SAM dataset mainly focuses on a static sound source, where the sound source position remains fixed within each sequence. Moreover, MOSPA requires the ground-truth absolute position of the sound source as a condition for motion generation, which is difficult to obtain in real-world scenarios and limits its applicability to embodied agents. Therefore, dynamic spatial-audio-driven human motion generation, which models human motion responses to time-varying spatial audio from moving sound sources, remains largely underexplored, especially in practical settings without access to ground-truth sound source positions.

To address this gap, we introduce \dbname, a dataset for dynamic spatial-audio-driven human motion generation. \dbname contains synchronized binaural audio, 3D human motion, and dynamic sound source trajectories captured in an indoor environment. Human motion and sound source trajectories are recorded using a Vicon motion capture system \citep{Vicon}, while binaural audio is captured by high-sensitivity microphones near the actors' ears, approximating the audio perceived by the actors. The main dataset contains 2,336 ten-second audio-motion sequences, totaling 23,360 seconds (approximately 6.5 hours), with motion at 30 FPS and audio at 48 kHz. It covers 42 audio types and 109 unique audio sources, including human sounds, animal sounds, vehicles, and environmental sounds, with 8 subjects and a binary genre label indicating sensitive or insensitive reactions. In addition, we provide \dbnamessl, an auxiliary sound source localization dataset of 11,730 seconds, pairing binaural audio of diverse acoustic events with listener head 6DoF poses and frame-wise moving sound source annotations to support dynamic spatial audio perception.

Motivated by the fact that human reactions to spatial sound depend on both where the sound comes from and what the sound is, we formulate spatial-audio-driven motion generation around three coupled functional components: \emph{localization}, \emph{understanding}, and \emph{generation}. Guided by this design principle, we propose \name, a diffusion-based framework for generating human motion conditioned on dynamic spatial audio. For localization, \name estimates frame-wise head-centric spatial cues of sound source from raw binaural audio using a sound source localization network, eliminating the reliance on privileged ground-truth source coordinates. For understanding, a pretrained CLAP audio encoder \citep{wu2023clap} encodes the binaural audio to obtain frame-wise audio-content features. For generation, \name employs a diffusion-based motion generator conditioned on the predicted spatial cues, CLAP audio features, and genre label. 
Since human motion needs to continuously adapt to the evolving spatial cues of the sound source, while self-attention only provides implicit interactions between motion tokens and condition tokens, we further introduce a \emph{Frame-wise Motion Correction} module in the Transformer denoiser to explicitly refine motion tokens with frame-wise conditions, along with a \emph{Correction Strength Predictor} that adaptively controls the correction magnitude for each frame.

By using head-centric DoA estimated directly from binaural audio, \name provides a more practical formulation for spatial-audio-driven motion generation in embodied robot settings. Compared with MOSPA~\citep{xu2025mospa}, which relies on ground-truth global sound source coordinates, our egocentric formulation avoids privileged source-location annotations and better reflects how embodied agents infer dynamic sound direction from their own auditory observations. Extensive quantitative and qualitative experiments show that \name outperforms existing baselines, producing more realistic motions that respond to dynamic spatial-audio cues.

Our contributions are summarized as follows:
\begin{itemize}[leftmargin=*,itemsep=0mm] 
    \item We introduce two datasets: \dbname, to the best of our knowledge the first dataset for dynamic spatial-audio-driven human motion generation, and \dbnamessl, an auxiliary sound source localization dataset collected in indoor reverberant environments.
    \item We propose \name{}, a practical framework for generating human motion from binaural audio without requiring ground-truth source coordinates. It integrates sound source localization, CLAP-based audio content encoding, and diffusion-based motion generation to model human responses to moving sound sources.
    \item Extensive quantitative and qualitative experiments show that \name generates natural, plausible, and spatially consistent human motion, better aligning with real motion and outperforming existing baselines.
\end{itemize}

%% file: sec/2_related.tex
\section{Related Work}
\vspace{-5pt}

\paragraph{Conditional Human Motion Generation.} Conditional human motion generation has been extensively studied under various conditions, such as action labels \citep{guo2020action2motion}, textual descriptions \citep{guo2022generating,zhang2023generating,dabral2023mofusion}, trajectories \citep{cai2026cmc}, and motion language models that treat human motion as a foreign language \citep{jiang2023motiongpt,zhang2024motiongpt}. These methods generate diverse and natural motions from semantic conditions, but they generally do not consider audio, let alone its spatial attributes. In contrast, our work focuses on audio-driven conditions with explicit spatial information.

\paragraph{Audio-Driven Human Motion Generation.} Audio-driven human motion generation has been studied mainly in two directions. The first is music-driven dance generation \citep{tang2018dance,zhuang2022music2dance,li2022danceformer,li2023finedance,siyao2022bailando,tseng2023edge,li2024lodge,luo2024popdg}, which generates choreography that matches the rhythm and style of the music. The second is speech-driven gesture synthesis \citep{ginosar2019learning,yoon2020speech,ao2022rhythmic,ao2023gesturediffuclip,alexanderson2023listen}, which generates co-speech gestures from speech content and prosody. Most of these methods use monaural or stereo audio and ignore the spatial attributes of the sound source. MOSPA \citep{xu2025mospa} is the first to study spatial-audio-driven human motion generation with binaural audio, but it assumes a static sound source position. Different from MOSPA, we consider dynamic spatial audio, where the source moves over time, and the DoA changes accordingly.

\paragraph{Spatial Sound Understanding.} Estimating the direction of a sound source from binaural signals is a classic problem in acoustic scene analysis, for instance using binaural cues combined with deep learning \citep{garcia2022binaural, min2025supervising}. 
Recent works also model room acoustics and room impulse responses to support sound source localization in complex indoor scenes \citep{chen2022soundspaces,luo2022learning,ratnarajah2022fast,um2025object, lan2024acoustic, lan2026smartdj}. 
In this paper, we adopt a sound source localization network to predict the head-centric DoA from binaural audio, which serves as the spatial condition for motion generation. This makes our framework practical in real-world scenarios where ground-truth source coordinates are unavailable.

%% file: sec/3_dataset.tex
\section{Datasets}

This section presents two newly collected motion-capture datasets for dynamic spatial-audio-driven human motion generation. 
The Dynamic Spatial Audio and Motion dataset (\dbname) serves as the main benchmark for this task, providing synchronized binaural audio, 3D human motions, and moving sound source trajectories. 
To support broader research on dynamic spatial audio perception, we introduce the Dynamic Spatial Audio Localization dataset (\dbnamessl) that pairs diverse acoustic events with listener head poses and mocap-recorded moving source trajectories.

\paragraph{Data Capture Setup.}

\begin{wrapfigure}{r}{0.5\textwidth}
    \centering
    \vspace{0mm}
    \includegraphics[width=\linewidth]{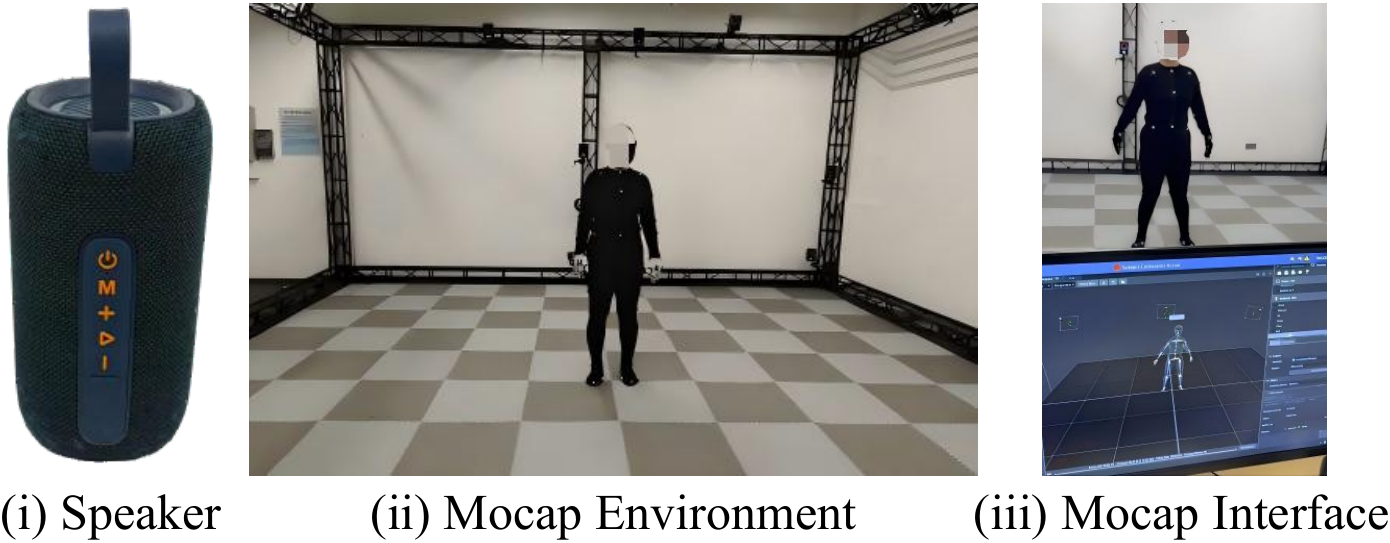}
    \vspace{-5mm}
    \caption{Data Collection Setup. }
    \vspace{0mm}
    \label{fig:teaser_small}
\end{wrapfigure}
We collect the data in an indoor environment with natural reverberation using a Vicon motion capture system~\citep{Vicon}, which records both human motion and the trajectory of a moving sound source (as shown in Fig.~\ref{fig:teaser_small}). We use an omnidirectional speaker to play the dynamic audio. We choose binaural audio as the audio modality, following the SAM dataset~\citep{xu2025mospa}, as it is a common form of spatial audio that aligns with human auditory perception and can be readily applied to virtual characters and robotic platforms. Binaural audio is recorded using two microphones placed near the actor's ears, allowing the signals to approximate the auditory input the actor perceives during capture. Specifically, we use Sound Professionals MS-TFB-2 microphones~\citep{mstfb2} and a Deity PR-2 recorder \citep{deity_pr2} for binaural audio recording. The audio recorder is synchronized with the mocap system using timecode, ensuring temporal alignment among binaural audio, human motion, and sound source trajectories.

\begin{table*}[t]
    
    \centering
    \caption{\textbf{Statistics of the \dbname dataset.} \dbname is the first dataset for dynamic spatial-audio-driven human motion generation, containing synchronized binaural audio, 3D human motion, and dynamic sound source trajectories (Dyn. SS). We also provide \dbnamessl, an auxiliary sound source localization dataset containing dynamic binaural recordings of diverse acoustic events with frame-wise source annotations.}
    \vspace{0mm}
    \resizebox{\textwidth}{!}{
    \begin{tabular}
    {lccccccc}
    \toprule
    Dataset & \textbf{Dyn. SS} & SS & 3D$\mathrm{Joint_{pos/rot}}$ & Model & Joints & Subjects & Hours \\
    \midrule
    Dance with Melody~\citep{tang2018dance} & $\times$ & $\times$ & \checkmark/$\times$ & - & 21 & - & 1.6 \\
    DanceNet~\citep{zhuang2022music2dance} & $\times$ & $\times$ & \checkmark/$\times$ & - & 55 & 2 & 0.9 \\
    AIST++~\citep{li2021ai} & $\times$ & $\times$ & \checkmark/\checkmark & COCO/SMPL & 17/24 & 30 & 5.2 \\
    PopDanceSet~\citep{luo2024popdg} & $\times$ & $\times$ & \checkmark/\checkmark & COCO/SMPL & 17/24 & 132 & 3.5 \\
    FineDance~\citep{li2023finedance} & $\times$ & $\times$ & \checkmark/\checkmark & SMPL \& hand joints & 52 & 27 & 14.6 \\
    SAM \citep{xu2025mospa} & $\times$ & \checkmark & \checkmark/\checkmark & SMPL-X & 55 & 12 & 9.5 \\

    \midrule

    \dbname (Ours) & \checkmark & \checkmark & \checkmark/\checkmark & SMPL-X & 55 & 8 & 6.5 \\
    + \dbnamessl (Ours) & \checkmark & \checkmark & \checkmark/\checkmark & SMPL-X & 55 & 1 & 3.3 \\

    \bottomrule
    \end{tabular}
    }
    \vspace{-2mm}
    \label{tab:dataset}
\end{table*}

\subsection{\dbname Dataset}

\begin{figure}[t]
    \centering

    \begin{subfigure}[b]{0.23\textwidth}
        \centering
        \includegraphics[width=\linewidth]{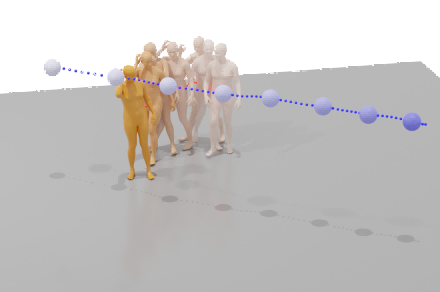}
        \caption{Siren}
    \end{subfigure}
    \hfill
    \begin{subfigure}[b]{0.23\textwidth}
        \centering
        \includegraphics[width=\linewidth]{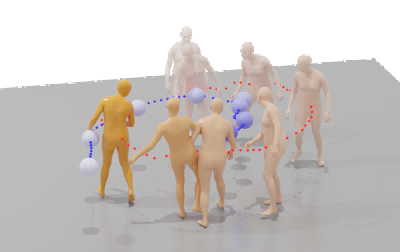}
        \caption{Dog barking}
    \end{subfigure}
    \hfill
    \begin{subfigure}[b]{0.23\textwidth}
        \centering
        \includegraphics[width=\linewidth]{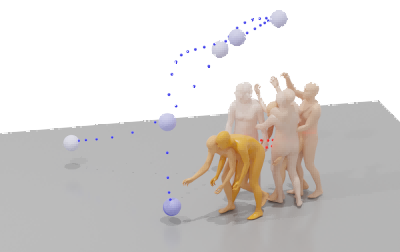}
        \caption{Birds chirping}
    \end{subfigure}
    \hfill
    \begin{subfigure}[b]{0.23\textwidth}
        \centering
        \includegraphics[width=\linewidth]{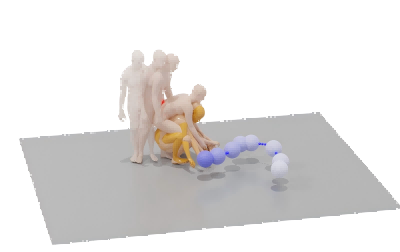}
        \caption{Cat meowing}
    \end{subfigure}

    \begin{subfigure}[b]{0.23\textwidth}
        \centering
        \includegraphics[width=\linewidth]{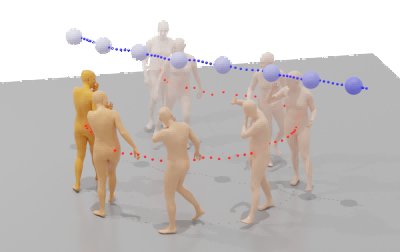}
        \caption{Person coughing}
    \end{subfigure}
    \hfill
    \begin{subfigure}[b]{0.23\textwidth}
        \centering
        \includegraphics[width=\linewidth]{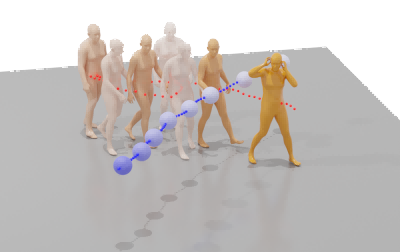}
        \caption{Car horn}
    \end{subfigure}
    \hfill
    \begin{subfigure}[b]{0.23\textwidth}
        \centering
        \includegraphics[width=\linewidth]{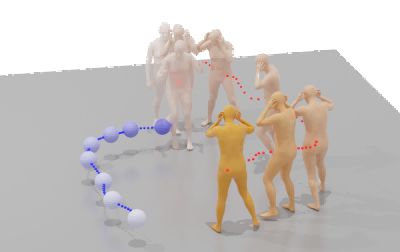}
        \caption{Fireworks exploding}
    \end{subfigure}
    \hfill
    \begin{subfigure}[b]{0.23\textwidth}
        \centering
        \includegraphics[width=\linewidth]{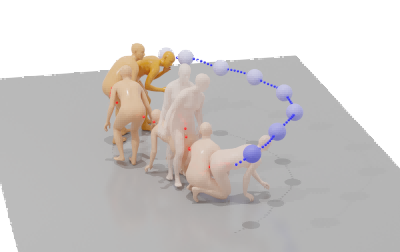}
        \caption{Handgun firing}
    \end{subfigure}

    \begin{subfigure}[b]{0.23\textwidth}
        \centering
        \includegraphics[width=\linewidth]{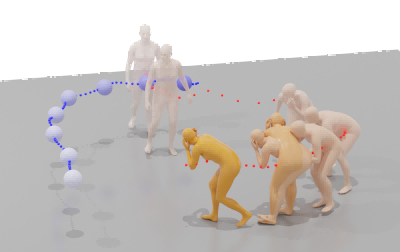}
        \caption{Machine gun firing}
    \end{subfigure}
    \hfill
    \begin{subfigure}[b]{0.23\textwidth}
        \centering
        \includegraphics[width=\linewidth]{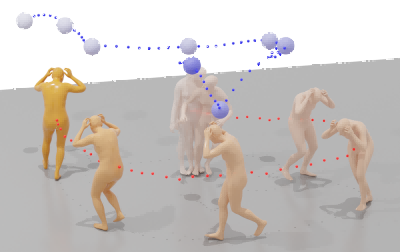}
        \caption{Bees buzzing}
    \end{subfigure}
    \hfill
    \begin{subfigure}[b]{0.23\textwidth}
        \centering
        \includegraphics[width=\linewidth]{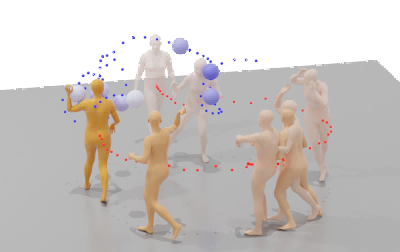}
        \caption{Bees buzzing}
    \end{subfigure}
    \hfill
    \begin{subfigure}[b]{0.23\textwidth}
        \centering
        \includegraphics[width=\linewidth]{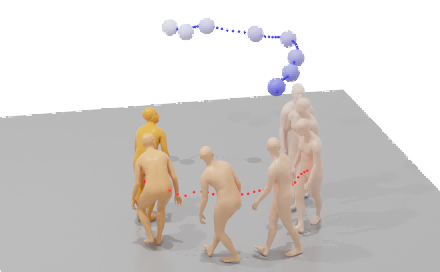}
        \caption{Monster roaring}
    \end{subfigure}

    \begin{subfigure}[b]{0.23\textwidth}
        \centering
        \includegraphics[width=\linewidth]{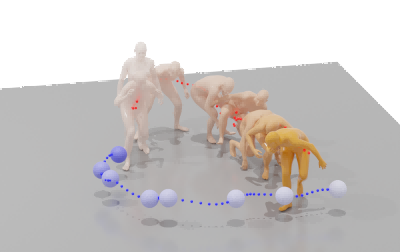}
        \caption{Monster footsteps}
    \end{subfigure}
    \hfill
    \begin{subfigure}[b]{0.23\textwidth}
        \centering
        \includegraphics[width=\linewidth]{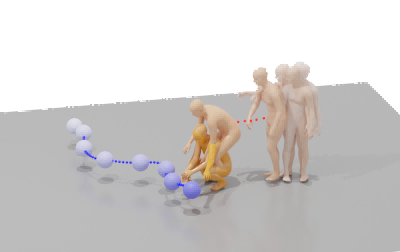}
        \caption{Puppy barking}
    \end{subfigure}
    \hfill
    \begin{subfigure}[b]{0.23\textwidth}
        \centering
        \includegraphics[width=\linewidth]{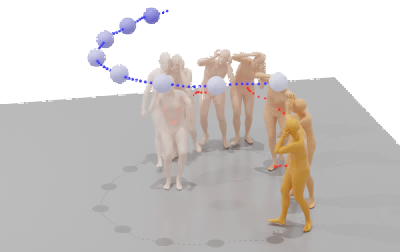}
        \caption{Rainstorm}
    \end{subfigure}
    \hfill
    \begin{subfigure}[b]{0.23\textwidth}
        \centering
        \includegraphics[width=\linewidth]{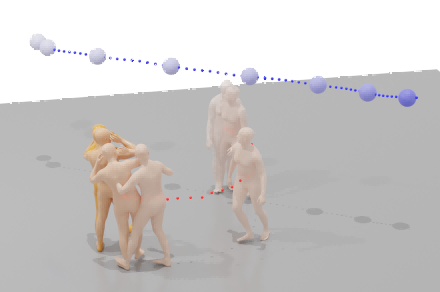}
        \caption{Train}
    \end{subfigure}

    \caption{Snapshots from the our \dbname dataset, spanning human sounds, animal sounds, and everyday environmental sounds. In each panel, the colors of the actor and the sound source transition from light to dark over time to indicate temporal evolution. Red dots mark the movement trajectory of the actor, while blue dots mark the trajectory of the moving sound source.}
    \label{fig:dataset_sound_sources}
\end{figure}

Our main dataset, \dbname, is collected for dynamic spatial-audio-driven human motion generation, with visualization snapshots provided in Fig.~\ref{fig:dataset_sound_sources}. Each sequence consists of a 10-second binaural audio clip, the corresponding 3D human motion stored in the SMPL-X format \citep{SMPL-X:2019}, and the trajectory of a moving sound source in the world coordinate system. As summarized in Tab.~\ref{tab:dataset}, \dbname contains 2,336 synchronized audio-motion sequences, amounting to 23,360 seconds (approximately 6.5 hours) of paired data. The captured motions are represented as SMPL-X model \citep{SMPL-X:2019}. The audio's sampling rate is 48,000\,Hz, while the motion is represented at 30\,FPS for network training. \dbname covers 42 audio types and 109 unique audio sources, spanning diverse daily-life acoustic events such as animal sounds, vehicles, alarms, music, and human speech. It was collected with 8 actors, including 5 female and 3 male. Each sequence is annotated with one of two reaction genres, sensitive or insensitive, with an approximately balanced 1:1 ratio.

\subsection{\dbnamessl Dataset}

We additionally collect \dbnamessl, an auxiliary sound source localization dataset, using the same motion capture setup. It provides abundant spatial supervision for training the SSL network, enabling robust estimation of dynamic head-centric DoA cues from binaural audio. It contains 11,730 seconds, approximately 3.3 hours, of binaural audio paired with frame-wise sound source positions and the listener's head 6DoF pose in the world coordinate system. These spatial cues can facilitate broader studies of spatial audio perception, such as absolute source localization and direction-of-arrival estimation. To ensure the diversity of acoustic events, we collect raw audio from AIST~\citep{aist-dance-db}, ESC-50~\citep{piczak2015esc}, and UrbanSound8K~\citep{salamon2014urbansound}, covering a wide range of music, environmental, human, animal, mechanical, and urban sounds.
Further details on the data collection and data processing pipeline for the two datasets are provided in Appendix~\ref{appendix:dataset}.

%% file: sec/4_method.tex
\section{Method} \label{sec:method}

\begin{figure}[t] 
    \vspace{0mm}
    \centering
       \includegraphics[width=\linewidth]{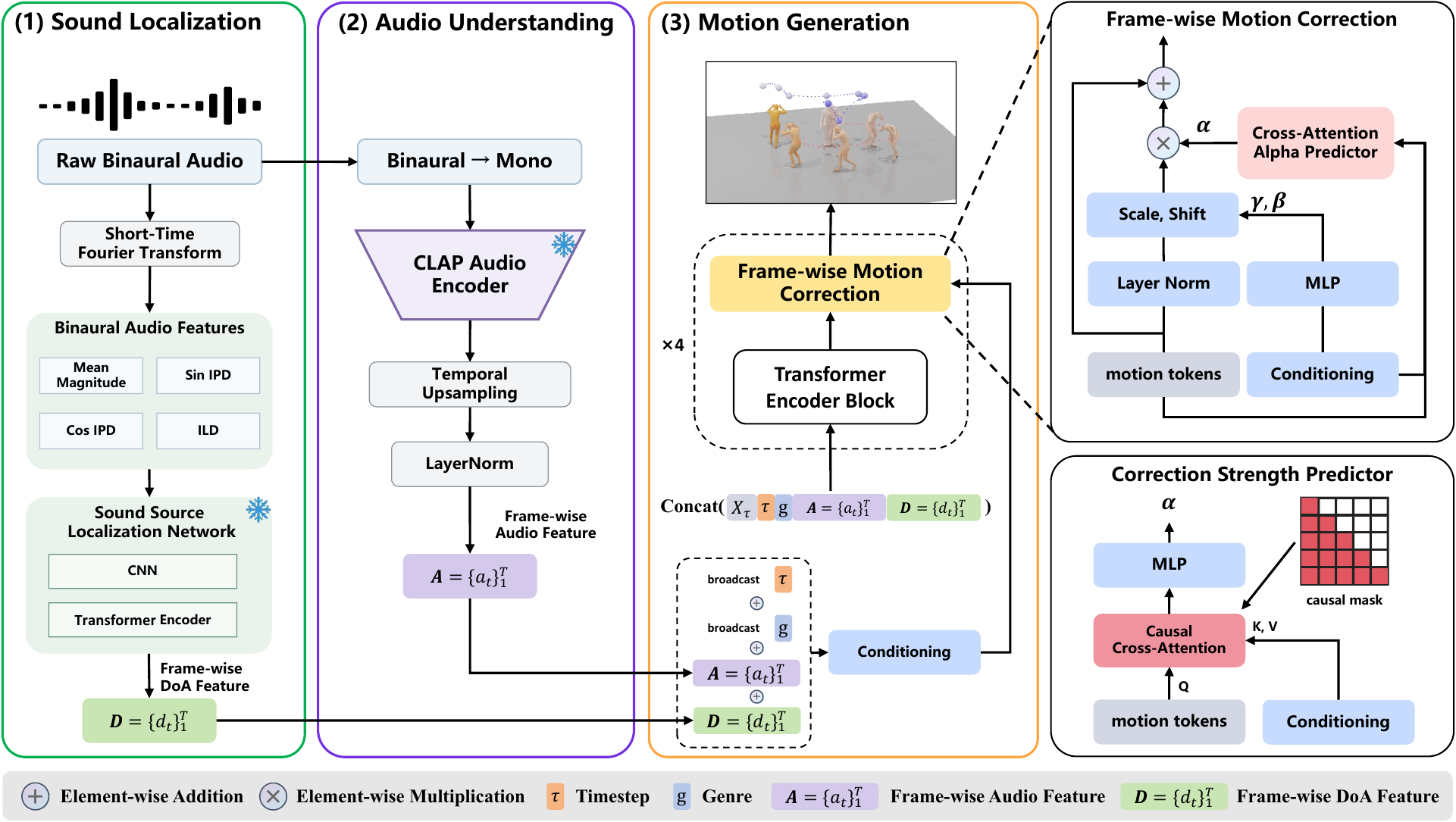}
       \caption{
            \textbf{Overview of \name framework.} \name{} comprises three coupled functional components: a sound source localization (SSL) network predicts a frame-wise head-centric DoA sequence $\hat{\boldsymbol{D}}$ from binaural audio (\emph{localization}); a CLAP audio encoder extracts frame-wise audio content features $\boldsymbol{A}$ from mono audio (\emph{understanding}); and a conditional diffusion model that iteratively denoise the Gaussian noise to generated motion, integrating Frame-wise Motion Correction to adaptively correct each motion according to frame-wise conditions (\emph{generation}).
            }
    \vspace{0mm}
    \label{fig:method}
\end{figure}

We propose \name{}, a diffusion-based framework for dynamic-spatial-audio-driven human motion generation. In real world, human reactions to spatial sound depend on both spatial direction and audio content.
Accordingly, \name{} consists of three coupled functional components: \emph{localization}, \emph{understanding}, and \emph{generation}, as illustrated in Fig.~\ref{fig:method}. For localization, a sound source localization (SSL) network estimates a frame-wise head-centric direction-of-arrival (DoA) sequence $\boldsymbol{D}=\{d_t \in \mathbb{R}^3\}_{t=1}^T$ from raw binaural audio, T is the max length of motion (Sec.~\ref{sec:ssl}). For understanding, a pretrained CLAP audio encoder extracts temporally aligned audio content representations $\boldsymbol{A}=\{a_t\}_{t=1}^T$ from the mono audio. For generation, a transformer-encoder-based conditional diffusion model synthesizes motion through frame-wise correction conditioned on $\boldsymbol{D}$, $\boldsymbol{A}$, the genre label $g$, and the diffusion timestep $\tau$. Rather than relying solely on attention mechanism, it incorporates \emph{Frame-wise Motion Correction} and \emph{Correction Strength Predictor}, allowing the denoiser to adaptively determine the correction magnitude for each frame under dynamic spatial-audio conditions (Sec.~\ref{sec:diffusion}). For network training, we adopt a 272-dimensional motion representation following \citep{xiao2025motionstreamer}.

\subsection{Sound Source Localization (SSL) Network} \label{sec:ssl}

In real-world scenes, ground-truth absolute global coordinates of sound sources are often unavailable. To obtain a motion-relevant and coordinate-invariant spatial cue, we represent the sound source at each frame as a head-centric direction-of-arrival (DoA) unit vector, transformed from the source position in the world coordinate system to the head coordinate system using the 6DoF head pose. Specifically, the SSL network predicts a sequence of frame-wise DoA vectors $\hat{\boldsymbol{D}}=\{\hat{d}_t\}_{t=1}^{T}$, where each $\hat{d}_t \in \mathbb{R}^3$ is a unit vector. This head-centric representation avoids the need for absolute global source coordinates.

Our SSL network is inspired by the binaural sound localization framework of Garcia et al.~\citep{garcia2022binaural}, which employs STFT-based binaural cues with a cascaded CNN-GRU architecture. We follow their use of STFT-based binaural spatial features and a CNN front-end for extracting local spectro-temporal patterns, but replace the GRU-based temporal modeling module with a Transformer encoder to better capture long-range temporal dependencies under reverberant environments. The SSL network is trained to predict the frame-wise head-centric DoA unit vector using a cosine-distance loss:
$
\mathcal{L}_{\mathrm{ssl}} = \frac{1}{T} \sum_{t=1}^{T} \left(1 - \hat{d}_t \cdot d_t\right),
$
where $\hat{d}_t$ and $d_t$ denote the predicted and ground-truth DoA unit vectors, respectively, both normalized in the head coordinate system. Once trained, the SSL network is frozen and used to predict the spatial condition sequence $\hat{\boldsymbol{D}}$. Please refer to Appendix \ref{appendix:ssl_net} for more details on the SSL architecture and training setup.

\subsection{Conditional Diffusion Motion Generation} \label{sec:diffusion}

We formulate motion generation as conditional denoising diffusion~\citep{ho2020denoising}. Given noisy motion $\mathbf{x}_{\tau}$ at diffusion step $\tau$, the denoising network predicts the clean motion sequence $\mathbf{x}_0$ conditioned on audio content, DoA, genre, and timestep embeddings. For audio content, we average the binaural signal into a mono waveform and feed it into a pretrained CLAP encoder~\citep{wu2023clap}. The extracted hidden features are pooled over channel and frequency dimensions and temporally aligned with the motion sequence, producing frame-wise audio features $\boldsymbol{A}=\{a_t\}_{t=1}^{T}$.

Building on the Transformer encoder of~\citet{xu2025mospa}, we project $\mathbf{x}_{\tau}$ into motion tokens $\boldsymbol{m}=\{\mathbf{m}_t\}_{t=1}^{T}$ and concatenate them with condition tokens derived from audio features, DoA features, genre, and the diffusion-step embedding. The concatenated tokens are fed into the Transformer encoder for conditional denoising. Since the attention mechanism only provides implicit interactions between motion tokens and the frame-wise condition tokens, we further introduce a \textit{Frame-wise Motion Correction} module to explicitly correct motion tokens based on frame-wise conditions, together with a \textit{Correction Strength Predictor} that adaptively determines the correction strength.

\noindent\textbf{Frame-wise Motion Correction.} 
To explicitly leverage both global and frame-wise conditions, we apply a residual correction module based on Adaptive Layer Normalization (AdaLN) on after each Transformer encoder layer to refine the intermediate motion token, which is formulated as
\begin{equation}
\boldsymbol{m}' = \boldsymbol{m} + \boldsymbol{\alpha} \odot \boldsymbol{\delta},
\end{equation}
where $\boldsymbol{\delta}=\{\delta_i\}_{i=1}^{T}$ denotes the frame-wise correction, and $\boldsymbol{\alpha}=\{\alpha_i\}_{i=1}^{T}$ is an frame-wise correction strength.
The modulation condition $\boldsymbol{c}=\{c_i\}_{i=1}^{T}$ consists of four components: the timestep embedding $t$, the genre embedding $g$, the frame-wise audio feature $\boldsymbol{A}$, and the frame-wise DoA feature $\boldsymbol{D}$. The unified condition sequence is obtained by summing these components at each frame, with the global timestep and genre embeddings broadcast to all $T$ frames: $ c_i = e_t + e_g + a_i + d_i$. This condition sequence is then used to modulate the motion tokens.
An MLP then estimates scale and shift sequences $(\boldsymbol{\gamma}, \boldsymbol{\beta}) = \mathrm{MLP}(\mathbf{c}_i) $ for AdaLN \citep{peebles2023dit}, yielding the correction:
\begin{equation}
\boldsymbol{\delta} =
\mathrm{FFN}\left(
\mathrm{LN}(\boldsymbol{m}) \odot (1 + \boldsymbol{\gamma}) + \boldsymbol{\beta}
\right),
\end{equation}

\noindent\textbf{Correction Strength Predictor.} 
The correction strength should not be a fixed parameter tied to a specific layer or frame position. Instead, the strength of motion correction should depend on the audio content, spatial DoA cues, genre, and the current denoising step. We therefore predict $\boldsymbol{\alpha}$ dynamically from the interaction between motion tokens and frame-wise conditions.

Specifically, we use motion tokens as queries $\boldsymbol{Q}=\boldsymbol{m}$ and modulation condition as keys and values $\boldsymbol{K}=\boldsymbol{V}=\boldsymbol{c}$. Considering that human responses to sounds may exhibit temporal delays, we apply a causal attention mask to prevent the token at frame $t$ from attending to future condition frames. The correct strength is bounded by
\begin{equation}
\boldsymbol{\alpha} = \alpha_{\max} \cdot \mathrm{Sigmoid}(
\mathrm{MLP}_{\alpha}(\mathrm{CrossAttn}(\boldsymbol{m}, \boldsymbol{c}_{\leq t}, \boldsymbol{c}_{\leq t}))
).
\end{equation}

This design decouples the estimation of the condition-dependent correction and its strength: the AdaLN modulation module predicts the correction term, while the cross-attention predictor dynamically determines how strongly this correction is applied.

\subsection{Training Objectives} \label{sec:loss}

We train \name a reconstruction loss:
$ \mathcal{L}_{\mathrm{diff}} = \left\| \boldsymbol{x_0} - \boldsymbol{\hat{x_0}}(\mathbf{x}_{\tau}, \tau, \mathbf{c}) \right\|_2^2. $
To ensure temporal continuity and reduce foot-sliding artifacts, we further use a delta loss
$
\mathcal{L}_{\mathrm{del}} =
\left\|
\hat{\boldsymbol{x_0}} - \boldsymbol{x_0}
\right\|_2^2
$
and a foot contact loss $\mathcal{L}_{\mathrm{foot}}$ \citep{xu2025mospa}. The overall objective is:
\begin{equation}
\mathcal{L} = \mathcal{L}_{\mathrm{diff}} + \lambda_{\mathrm{del}}\mathcal{L}_{\mathrm{del}} + \lambda_{\mathrm{foot}}\mathcal{L}_{\mathrm{foot}}.
\end{equation}

%% file: sec/5_experiment.tex
\begin{figure}[t] 
    \vspace{-5mm}
    \centering
       \includegraphics[width=\linewidth]{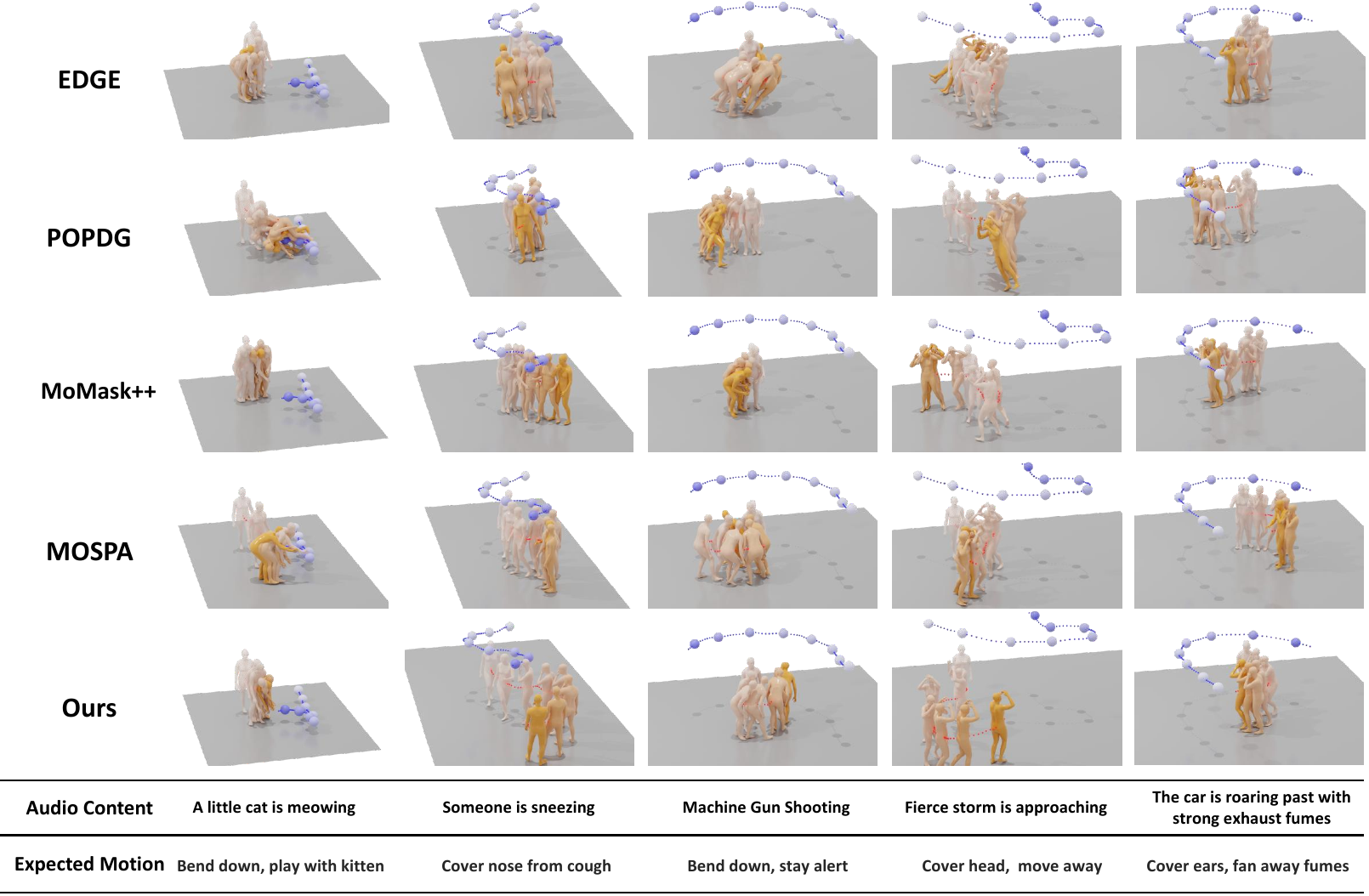}
       \caption{
            Qualitative comparison of state-of-the-art methods for the dynamic spatial audio-to-motion task. We visualize motion results from five cases.
            }
    \vspace{0mm}
    \label{fig:qualitative}
\end{figure}

\section{Experiment}

\paragraph{Experimental Setup.}
We split the main \dbname dataset into train/validation/test sets with a ratio of 8:1:1. For training the sound source localization network, we transform the global sound source position into head-centric unit direction vector using the head 6DoF pose, and use it as the ground-truth DoA. Regarding metrics, we use FID, R-precision, Diversity~\citep{guo2022generating}, Average Pairwise Distance (APD)~\citep{dou2023c}, and a physics-based skating ratio, as detailed in Appendix \ref{appendix:metrics}. We introduce a multi-branch motion-audio-source evaluator for feature extraction, aiming to better measure the spatial alignment between the generated motion and the moving sound source; details are provided in Appendix \ref{appendix:evaluator}.
We compare with diffusion-based and VQ-based baselines, including EDGE~\citep{tseng2023edge}, POPDG~\citep{luo2024popdg}, MOSPA~\citep{xu2025mospa}, and MoMask++~\citep{hwang2025snapmogen}. All baselines are trained on the same \dbname split, with their original input conditions replaced by traditional acoustic features for fair comparison.

\subsection{Comparisons}

Table~\ref{tab:main} reports the quantitative comparisons on the \dbname{} test set. \name{} achieves best overall performance across all metrics. Among the baselines, MoMask++~\citep{hwang2025snapmogen} obtains a competitive FID but suffers from lower R-precision and a higher skating ratio, indicating weaker audio-motion alignment and physical plausibility.
MOSPA~\citep{xu2025mospa} achieves the second-best R-precision by using ground-truth source positions, while \name{} estimates head-centric DoA directly from binaural audio and obtains better overall results. Qualitative comparisons are shown in Fig.~\ref{fig:qualitative}, with more results in the supplementary videos.

\begin{table}
    \centering
    \footnotesize
    \caption{\textbf{Quantitative Comparisons on the \dbname Dataset.} ± indicates a 95\% confidence interval. \textbf{Bold face} indicates the best result. $\rightarrow$ means the closer to Real Motion the better.}
    \vspace{0mm}
    \setlength{\tabcolsep}{0.5pt}
    \resizebox{\linewidth}{!}{
    \begin{tabular}{ccccccccc}
        \toprule
        \multirow{2}{*}{Method} 
        & \multicolumn{3}{c}{$\text{R-precision}\uparrow$}
        & \multirow{2}{*}{$\text{FID}\downarrow$}
        & \multirow{2}{*}{$\text{Diversity} \rightarrow$}
        & \multirow{2}{*}{$\text{APD} \rightarrow$}
        & \multirow{2}{*}{$\text{Skating Ratio (\%)}\downarrow$} \\

        \cmidrule{2-4}
        & Top1 & Top2 & Top3 & & & & \\
        \midrule

        Real Motion
        & \et{0.720}{0.000} & \et{0.886}{0.000} & \et{0.976}{0.000}
        &
        $0.001$ &
        \et{29.66}{0.512} &
        45.53 &
        $1.26$ \\

        \midrule

        EDGE~\citep{tseng2023edge}
        & \et{0.457}{0.006} & \et{0.730}{0.010} & \et{0.819}{0.009} & \et{7.498}{0.485} & \et{27.49}{0.119} & 35.44 & 9.23 \\

        POPDG~\citep{luo2024popdg} &
        \et{0.462}{0.006} &
        \et{0.681}{0.005} &
        \et{0.782}{0.004} &
        \et{5.727}{0.485} &
        \et{28.15}{0.141} &
        34.99 &
        2.77 \\

        MOSPA \citep{xu2025mospa}
        & \et{0.585}{0.013} & \et{0.788}{0.009} & \et{0.885}{0.006} & \et{5.160}{0.875}
        & \et{33.50}{0.583}
        & 50.29
        & 3.69 \\

        MoMask++ \citep{hwang2025snapmogen}
        & \et{0.415}{0.006} & \et{0.656}{0.008} & \et{0.775}{0.009} & \et{3.909}{0.331} & \et{28.39}{0.787} & 38.24 & 14.11 \\

        \name (Ours)
        & \etb{0.623}{0.004} & \etb{0.824}{0.003} & \etb{0.920}{0.004} & \etb{0.857}{0.082} 
        & \etb{28.97}{0.636}
        & \textbf{42.60}
        & \textbf{1.66} \\
      \bottomrule
    \end{tabular}
    }
    \vspace{0mm}
    \label{tab:main}
\end{table}

\begin{table}[t]
    \centering
    \vspace{0mm}
    \Large
    \caption{\textbf{Ablation on the \dbname{} Dataset,} including the types of sound source and audio features, Frame-wise Correction (Frame Corr.) , and Correction Strength Predictor (Strength Pred.).}
    \renewcommand{\arraystretch}{1.1}
    \resizebox{\textwidth}{!}{
    \begin{tabular}[t]{c | c | c | c c c c | c c c c c}
    \toprule
        \multirow{2}{*}{Item}
        & \multirow{2}{*}{\shortstack{Ablated \\ Component}}
        & \multirow{2}{*}{Method}
        & \multirow{2}{*}{\shortstack{Sound \\ Source}}
        & \multirow{2}{*}{\shortstack{Audio \\ Feature}}
        
        & \multirow{2}{*}{\shortstack{Frame \\ Corr.}}
        & \multirow{2}{*}{\shortstack{Strength \\ Pred.}}
        & \multirow{2}{*}{FID $\downarrow$}
        & \multicolumn{3}{c}{R-precision $\uparrow$}
        & \multirow{2}{*}{\parbox{2.0cm}{\centering Skating $\downarrow$ \\ Ratio (\%)}} \\
    \cline{9-11}

        & & & & & & & & Top1 & Top2 & Top3 & \\
    \midrule

    &
    & Real Motion
    &
    &
    &
    &
    &
    & \et{0.720}{0.000} & \et{0.886}{0.000} & \et{0.976}{0.000}
    & 1.26 \\

    \midrule

    \circled{1}
    & \multirow{3}{*}{Sound Source}
    & w/o Sound Source
    & $\emptyset$
    & Frame-wise CLAP
    & \Checkmark
    & \Checkmark
    & \et{1.334}{0.236} & \et{0.359}{0.010} & \et{0.593}{0.021} & \et{0.722}{0.011}
    & 2.25 \\

    \circled{2}
    &
    & World Position
    & Position
    & Frame-wise CLAP
    & \Checkmark
    & \Checkmark
    & \et{2.222}{0.318} & \et{0.549}{0.012} & \et{0.790}{0.008} & \et{0.895}{0.005}
    & 2.77 \\

    \circled{3}
    &
    & GT DoA
    & GT DoA
    & Frame-wise CLAP
    & \Checkmark
    & \Checkmark
    & \etb{0.726}{0.056} & \etb{0.673}{0.006} & \et{0.839}{0.003} & \etb{0.936}{0.001}
    & 2.32 \\

    \midrule

    \circled{4}
    & \multirow{2}{*}{Audio Feature}
    & Trad. Feat.
    & Pred DoA
    & Trad. Feat.
    & \Checkmark
    & \Checkmark
    & \et{2.343}{0.160} & \et{0.547}{0.016} & \et{0.768}{0.011} & \et{0.888}{0.011}
    & 4.68 \\

    \circled{5}
    &
    & Global CLAP
    & Pred DoA
    & Global CLAP
    & \Checkmark
    & \Checkmark
    & \et{6.693}{0.726} & \et{0.569}{0.012} & \et{0.767}{0.007} & \et{0.860}{0.011}
    & 3.28 \\

    \midrule

    \circled{6}
    & \multirow{3}{*}{Network Module}
    & w/o Corr.
    & Pred DoA
    & Frame-wise CLAP
    &
    &
    & \et{4.458}{0.328} & \et{0.556}{0.011} & \et{0.772}{0.010} & \et{0.896}{0.006}
    & 3.71 \\

    \circled{7}
    &
    & w/o Strength Pred.
    & Pred DoA
    & Frame-wise CLAP
    & \Checkmark
    &
    & \et{1.530}{0.185} & \et{0.586}{0.009} & \et{0.813}{0.004} & \et{0.917}{0.002}
    & 2.41 \\

    \circled{8}
    & 
    & Trans. Dec.
    & Pred DoA
    & Frame-wise CLAP
    & \Checkmark
    & \Checkmark
    & \et{0.906}{0.138} & \et{0.627}{0.008} & \etb{0.853}{0.005} & \et{0.915}{0.004}
    & 1.81 \\
    
    \midrule

    \circled{9}
    &
    & \textbf{Ours (full model)}
    & Pred DoA
    & Frame-wise CLAP
    & \Checkmark
    & \Checkmark
    & \et{0.857}{0.082} & \et{0.623}{0.005} & \et{0.824}{0.003} & \et{0.920}{0.004}
    & \textbf{1.66} \\

    \bottomrule
    \end{tabular}
    }
    \vspace{-4mm}
    \label{tab:ablation2}
\end{table}

\subsection{Ablation Study}

We conduct several ablation experiments to evaluate each component's contribution, including the type of sound source, audio feature and network module. Each variant is trained with the totally same training schedule, and the results are summarized in Tab.~\ref{tab:ablation2}.
\vspace{-4mm}
\paragraph{Sound Source.}
This group evaluates the effect of sound source. 
Replacing the predicted DoA with source positions (\circled{2}) leads to a clear performance drop. 
Removing the sound source condition entirely (\circled{1}) leads to a drastic decrease in R-precision, highlighting the importance of spatial cues for audio-motion alignment.
As expected, using the ground-truth DoA (\circled{3}) achieves the best performance, confirming the importance of accurate head-centric spatial cues for generating spatially and semantically aligned motions.
\vspace{-4mm}
\paragraph{Audio Feature.}
This group studies the effect of different audio representations by replacing the frame-wise CLAP embedding in the full model with either hand-crafted acoustic features (\circled{4}) or a global CLAP feature pooled over the entire audio segment (\circled{5}). Both alternatives lead to obvious degraded performance. This suggests that frame-wise CLAP features are important for providing better representation for audio content, which are crucial for generating high-quality motions.
\vspace{-4mm}
\paragraph{Network Module.}
This group investigates our proposed conditioning modules. Removing the \emph{Correction Strength Predictor} leads to degradation in all metrics, indicating that adaptively estimating the correction strength is important for both motion quality and audio-motion alignment (\circled{7}).
When the \emph{Frame-wise Residual Correction}  is further removed (\circled{6}), the performance drops more significantly, suggesting that simply concatenating condition tokens and relying on self-attention cannot explicitly align motion with frame-wise conditions. We further ablate the main backbone by replacing it with a Transformer decoder (\circled{8}), which leads to a clear performance drop.



\begin{wraptable}{r}{0.47\textwidth}
    \centering
    \vspace{-4mm}
    \caption{\textbf{Effect of the auxiliary \dbnamessl training data.}
    We report the angular error on the \dbnamessl validation set.}
    \renewcommand{\arraystretch}{1.1}
    \resizebox{\linewidth}{!}{
    \begin{tabular}{l | c}
        \toprule
        SSL Training Data
        & \shortstack{\dbnamessl Val. Angular Error
          (\ensuremath{{}^{\circ}}) \ensuremath{\downarrow}} \\
        \midrule
        DySAM
        & 48.92 \\
        DySAM + \dbnamessl{}
        & \textbf{36.62} \\
        \bottomrule
    \end{tabular}
    }
    \label{tab:dysal_training_data}
    \vspace{-3mm}
\end{wraptable}
\paragraph{Ablation Study on Sound Source Localization Network.}
We evaluate the architectural design of the SSL network, focusing on the DoA representation and temporal modeling module. All variants follow the same training schedule and are evaluated by frame-wise angular error on the \dbname{} validation set. As shown in Table~\ref{tab:ssl_ablation}, unit direction vectors outperform spherical coordinates, while the Transformer encoder further improves localization accuracy over Bi-GRU, supporting its effectiveness in temporal spatial-audio modeling. In addition, Table~\ref{tab:dysal_training_data} shows that incorporating the \dbnamessl{} training split reduces the angular error on the \dbnamessl{} validation set, indicating that its frame-wise annotations provide useful spatial supervision for the diverse acoustic events covered by the dataset. The SSL network used in \name{} adopts the unit-direction-vector representation, the Transformer encoder, and joint training on \dbname{} and \dbnamessl{}.

\begin{table}[t]
    \centering
    \vspace{2mm}
    \caption{\textbf{Ablation of the sound source localization network.}
    We evaluate the DoA representation and temporal architecture on the
    DySAM validation set. All variants are trained using the same training
    data. Reported numbers are the full-frame angular errors
    (\ensuremath{{}^{\circ}}) between the predicted and ground-truth
    source directions.}
    \renewcommand{\arraystretch}{1.1}
    \resizebox{0.93\textwidth}{!}{
    \begin{tabular}{c | c | c c | c}
        \toprule
        & Method
        & \multirow{2}{*}{\shortstack{DoA \\ Representation}}
        & \multirow{2}{*}{\shortstack{Temporal \\ Module}}
        & \multirow{2}{*}{\shortstack{DySAM Val. \\ Angular Error
          (\ensuremath{{}^{\circ}}) \ensuremath{\downarrow}}} \\
        & & & & \\
        \midrule

        \circled{1}
        & Using Spherical
        & Spherical (azimuth, elevation)
        & Bi-GRU
        & 18.52 \\

        \circled{2}
        & Using GRU
        & Unit direction vector
        & Bi-GRU
        & 16.76 \\

        \circled{3}
        & Using Transformer
        & Unit direction vector
        & Transformer encoder
        & \textbf{12.87} \\

        \bottomrule
    \end{tabular}
    }
    \vspace{-2mm}
    \label{tab:ssl_ablation}
\end{table}

%% file: sec/6_conclusion.tex
\section{Conclusion}

This paper introduces the task of dynamic spatial-audio-driven human motion generation. We present \dbname{}, the first dataset and benchmark for this task containing synchronized binaural audio, 3D human motion, and moving sound source trajectories, together with \dbnamessl{}, an auxiliary dataset that provides diverse acoustic events and frame-wise spatial annotations for dynamic spatial audio perception.
We further propose \name{}, a diffusion-based framework that explicitly leverages frame-wise audio-spatial conditions to generate high-quality human motions in response to dynamic spatial audio.
Extensive experiments demonstrate that \name{} produces natural and physically plausible motions with improved alignment to dynamic spatial audio.
\textit{Limitations and Future Work:} Our current setting considers a single prominent moving sound source. Extending the framework to multiple simultaneous sources would require jointly reasoning about source separation, localization, and their potentially different behavioral relevance. We leave this challenging setting for future research and evaluation.

%% file: sec/7_appendix.tex
\clearpage
\appendix
\renewcommand\thefigure{\Alph{section}\arabic{figure}}    
\renewcommand\thetable{\Alph{section}\arabic{table}}
\setcounter{figure}{0}
\setcounter{table}{0}

\section{Dataset Details}
\label{appendix:dataset}

\subsection{Motion Capture Design}

At the beginning of each clip, the dynamic sound source, i.e., a speaker, starts from one of four azimuthal regions relative to the actor: front-left, rear-left, rear-right, and front-right. This design enables us to capture diverse human reactions to sounds arriving from different directions. The trajectory of the sound source is designed to be either straight or curved. Moreover, the sound source either moves toward the actor and reaches the actor's position, or passes by without contact. These two movement patterns elicit fundamentally different responses. For example, an actor may react differently when a dog or a car rushes toward them than when it merely passes by.

We further define two reaction styles: insensitive and sensitive, to represent different levels of responsiveness to the same acoustic event. For instance, when hearing an explosion, an insensitive actor may remain largely unreactive, whereas a sensitive actor may immediately move away from the sound source. The two reaction styles are approximately balanced in the dataset.

\subsection{Motion Capture and Audio Recording Setup}

The data are collected in a typical indoor environment with natural room reverberation. The motion capture system consists of 20 Vicon cameras installed around the capture space, recording both the human body motion and the sound source trajectory at 120\,Hz.

For the moving sound source, we use a cylindrical omni-directional speaker to play the audio, ensuring stable sound radiation in different directions as the speaker moves. For binaural audio recording, a high-sensitivity microphone is placed near each ear of the actor to capture the audio perceived by the actor as faithfully as possible. The two microphones are connected to a Deity PR-2 recorder~\citep{deity_pr2}, which is synchronized with the Vicon system in advance using timecode. This setup ensures accurate temporal alignment between the recorded binaural audio and the corresponding motion capture data.

\subsection{Spatial Audio Scenarios in \dbname}

The \dbname dataset covers 42 distinct audio types, corresponding to 109 unique audio sources, as listed in Tab. \ref{tab:acoustic_scene}. Each audio type comprises one to six audio sources, and for each audio source 8 to 32 ten-second motion sequences are collected, yielding 2{,}336 (binaural audio, motion) pairs in total. These audio types span daily-life sounds including animal vocalizations, vehicles, environmental noises, alarms, musical instruments, and human speech.

\begin{table}[h]
    \centering
    \small
    \setlength{\tabcolsep}{6pt}
    \renewcommand{\arraystretch}{1}
    \begin{tabular}{c  c  c  c}
        \hline
        \multicolumn{4}{c}{\textbf{Audio Type}} \\
        \hline
        alarm                & bark               & bicycle\_bell      & bird \\
        car\_bidirect        & car\_fast          & car\_slow          & cat \\
        cough                & drill              & drill\_laugh       & engine\_slow \\
        firework             & gun\_hand          & gun\_machine       & helicopter \\
        horn\_car            & horn\_train        & horn\_truck        & insect\_bees \\
        insect\_flying       & insect\_msqt       & monster\_roar      & monster\_step \\
        motor\_fast          & motor\_slow        & plane              & puppy \\
        rainstorm            & rockroll           & sneezing           & train \\
        violin               & voice\_danger\_female & voice\_danger\_male & voice\_follow\_female \\
        voice\_follow\_male  & voice\_greet\_female & voice\_greet\_male   & voice\_help\_female \\
        voice\_help\_male    & witch              &                    &  \\
        \hline
    \end{tabular}
    \vspace{2mm}
    \caption{The 42 distinct audio types covered in \dbname, corresponding to 109 unique audio sources. Each audio source is accompanied by 8 to 32 ten-second motion sequences, and the dataset contains 2{,}336 (binaural audio, motion) pairs in total.}
    \label{tab:acoustic_scene}
\end{table}

\subsection{Audio Sources for \dbnamessl}

To ensure broad audio diversity, we collect raw audio materials from AIST~\citep{aist-dance-db}, ESC-50~\citep{piczak2015esc}, and UrbanSound8K~\citep{salamon2014urbansound}. AIST mainly contains music and dance-related audio with diverse rhythms, styles, and structures. ESC-50 provides environmental sounds from natural scenes, human activities, animal sounds, indoor and outdoor daily scenes, and mechanical devices, such as rain, sea waves, bird calls, coughing, footsteps, keyboard typing, bells, and vacuum cleaners. UrbanSound8K contains typical urban sounds, including air conditioners, car horns, children playing, dog barks, drilling, engine idling, gunshots, jackhammers, sirens, and street music.

By combining these sources, \dbnamessl{} covers music, human activities, natural environments, animals, indoor and outdoor daily activities, transportation, mechanical construction, alarms, and urban public spaces. This diversity broadens the range of acoustic events available for training and evaluating sound source localization models under dynamic spatial-audio conditions.

\subsection{Data Processing}

\begin{figure}[t]
    \centering

    \begin{subfigure}[t]{0.48\textwidth}
        \centering
        \includegraphics[width=\linewidth]{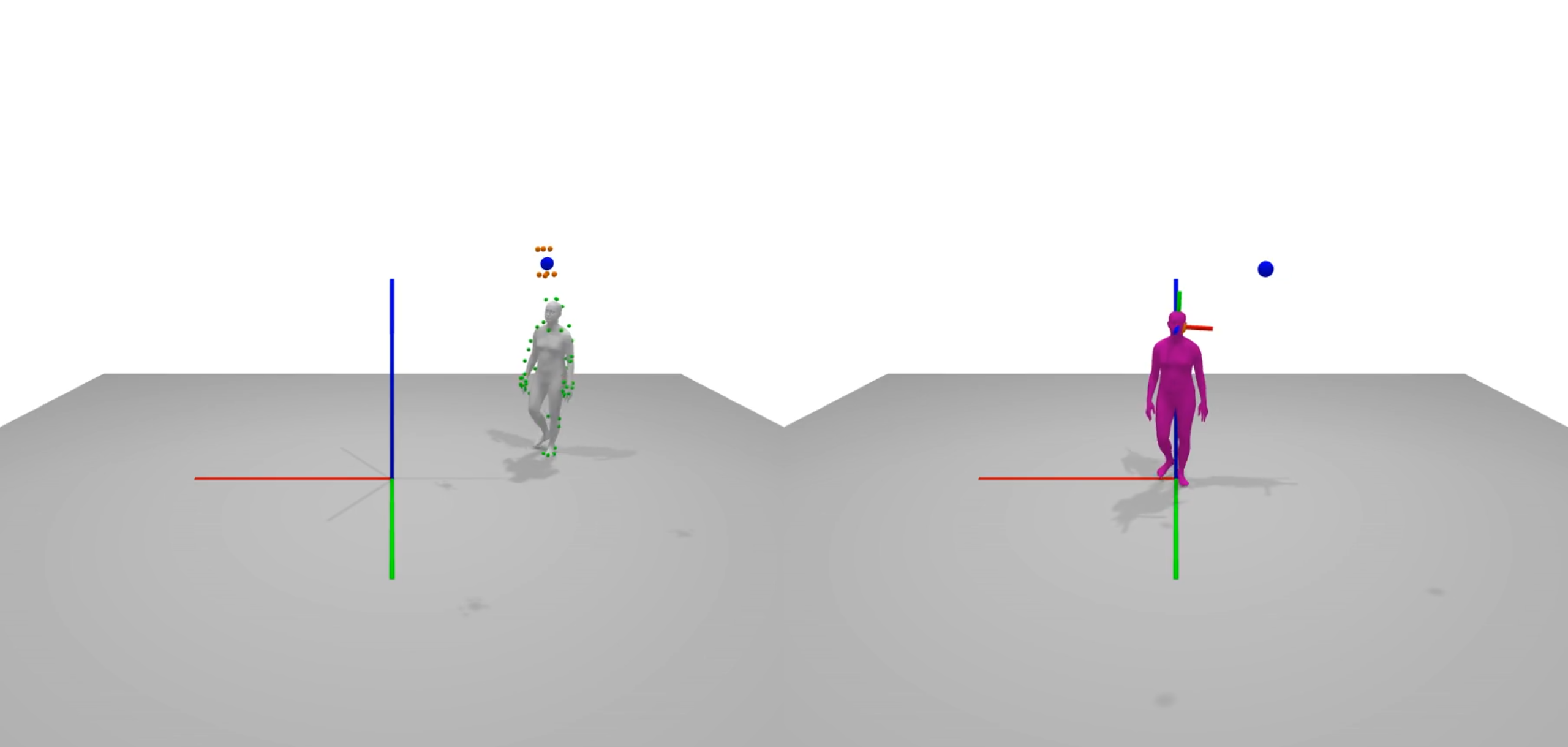}
        \caption{Example 1}
    \end{subfigure}
    \hfill
    \begin{subfigure}[t]{0.48\textwidth}
        \centering
        \includegraphics[width=\linewidth]{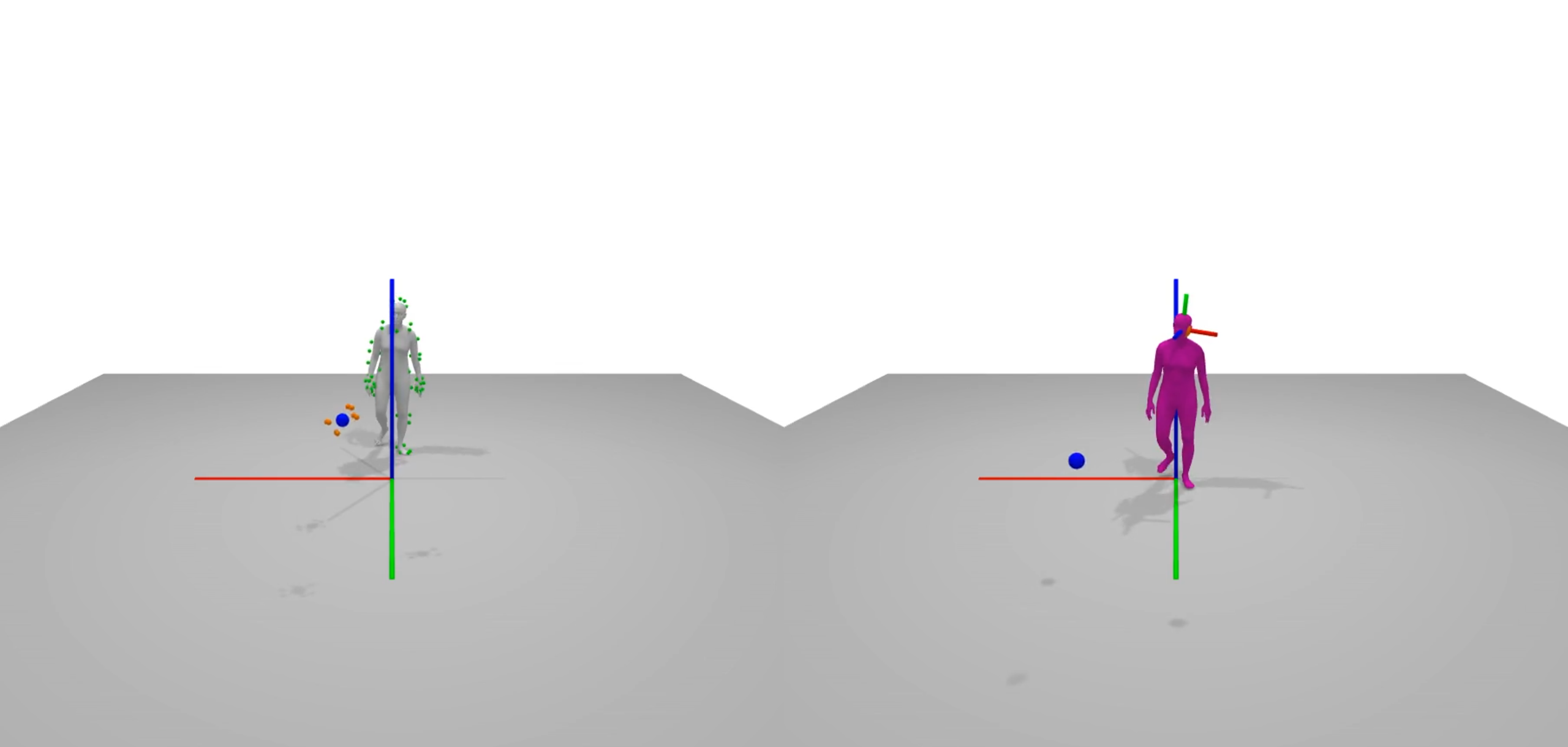}
        \caption{Example 2}
    \end{subfigure}

    \begin{subfigure}[t]{0.48\textwidth}
        \centering
        \includegraphics[width=\linewidth]{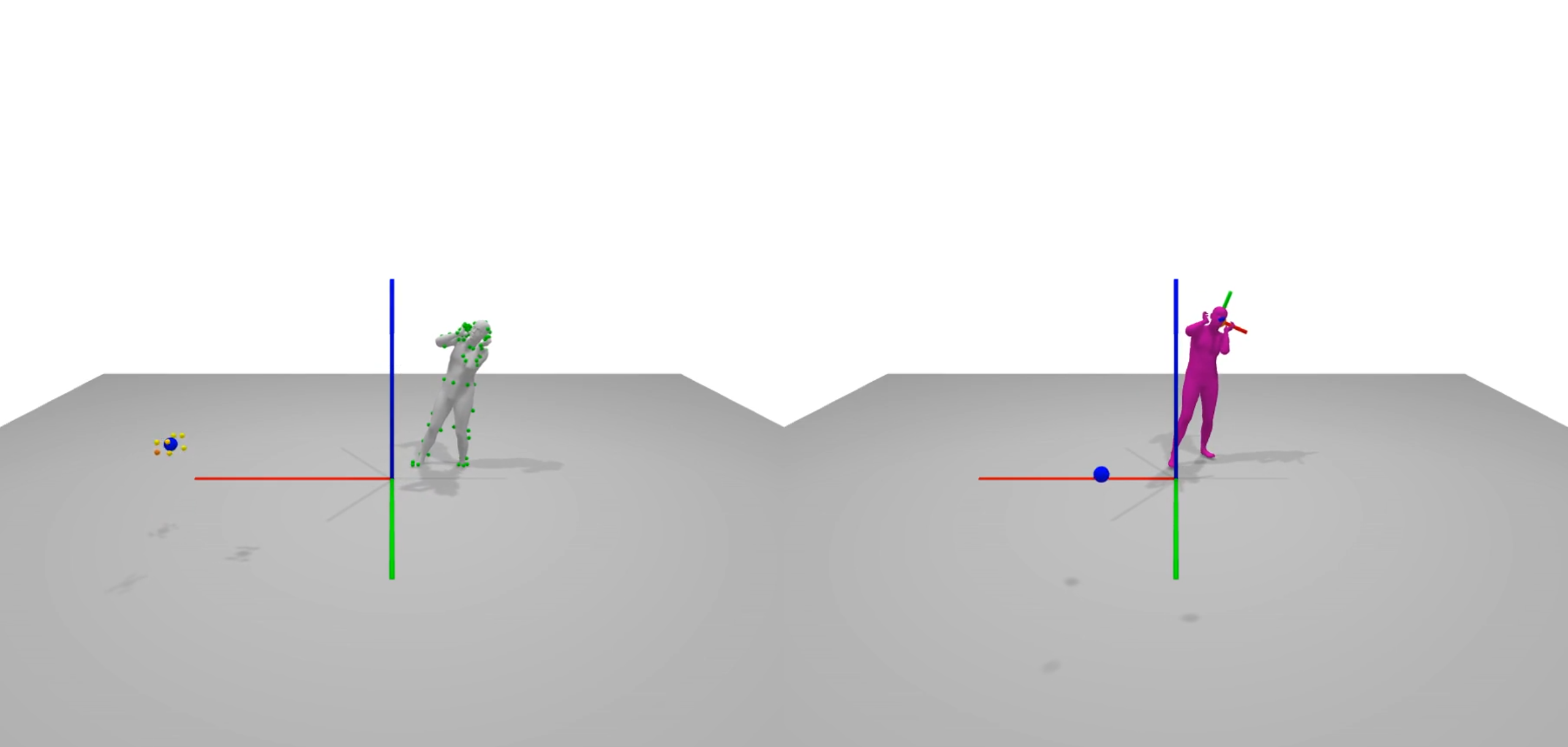}
        \caption{Example 3}
    \end{subfigure}
    \hfill
    \begin{subfigure}[t]{0.48\textwidth}
        \centering
        \includegraphics[width=\linewidth]{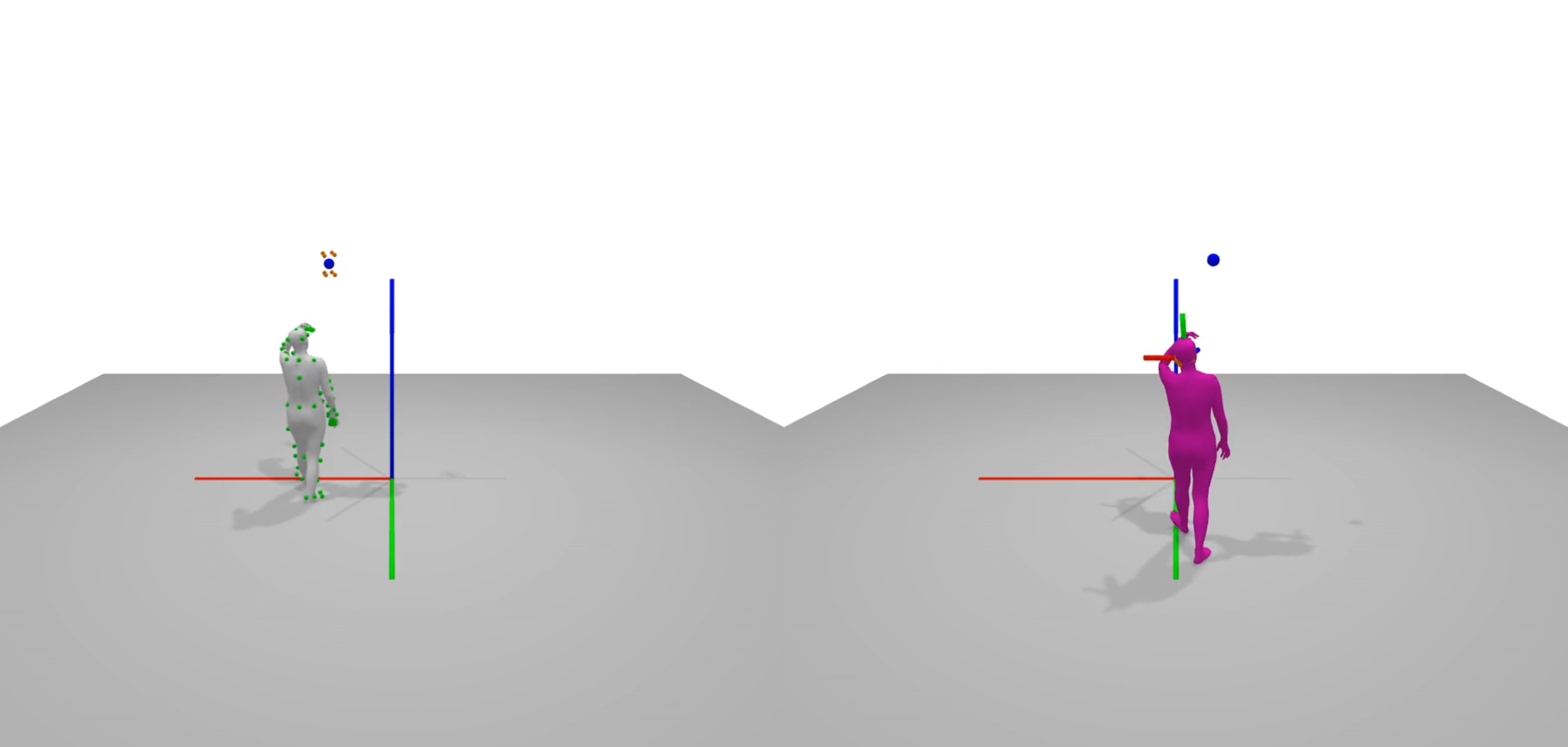}
        \caption{Example 4}
    \end{subfigure}

    \caption{Four examples of \dbname dataset. In each image, the left column shows the original C3D body markers and speaker markers, SMPL-X motion \textcolor{gray}{(gray body)}, and calculated speaker location; the right column shows the canonicalized SMPL-X motion \textcolor{purple}{(purple body)} and the head coordinate system. The red, green, and blue axes denote the x, y, and z axes, respectively.}
    \label{fig:c3d_smplx_example}
\end{figure}

For intuitively understand our data processing pipeline, we provide four visualization examples of the \dbname dataset in Fig.~\ref{fig:c3d_smplx_example}.

\textbf{Motion processing.}
During capture, the actor wears a motion capture suit with multiple reflective markers attached to the body, represented by the \textcolor{green}{green points} in Fig.~\ref{fig:c3d_smplx_example} The Vicon system exports the captured data as C3D files, where each file stores the world coordinates of all body markers for every frame. We use Mosh++ \citep{mahmood2019amass} to fit each C3D sequence into the SMPL-X parameters. The fitted SMPL-X motion is then canonicalized by rotating the human body in the initial frame to face the positive $y$-axis and move the center of the body to the origin. For training the motion generation model, we further convert the SMPL-X motion into the 272-dimensional motion representation following~\citep{xiao2025motionstreamer}.

\textbf{Sound source location processing.}
To track the position of the sound source, i.e., the speaker, we attach eight reflective markers around the speaker (\textcolor{orange}{orange points} in Fig.~\ref{fig:c3d_smplx_example}), which approximately correspond to the eight vertices of a cuboid. For each C3D sequence, we use the speaker markers in the initial frame as reference markers. We then estimate the speaker translation in each subsequent frame by solving the rigid alignment between the reference markers and the current-frame markers using singular value decomposition (SVD). The estimated translation is used as the global sound source location in our dataset,  represented by the \textcolor{blue}{blue sphere} in Fig.~\ref{fig:c3d_smplx_example}. In practice, some speaker markers may be missing in certain frames due to occlusion or tracking failure, making SVD-based estimation unreliable or infeasible. For these frames, we use Kalman filtering to smoothly interpolate and complete the missing sound source locations.

\textbf{Head-centric DoA processing.}
To train the sound source localization network, we further compute the head-centric sound source location. Specifically, for each frame, we perform forward kinematics on the fitted SMPL-X body model to obtain the translation and rotation of the head joint, i.e., joint 15. We then transform the global sound source location into the head coordinate system using the estimated head pose, yielding the head-centric sound source location as the direction-related supervision signal.

\section{Implementation Details of the SSL Network} \label{appendix:ssl_net}

We provide additional implementation details of the sound source localization network used in Sec.~\ref{sec:ssl}. The SSL network estimates a head-centric direction of arrival from binaural audio. For every frame, it predicts a unit vector
$
\mathbf{d}_t \in \mathbb{R}^{3},
$
which points from the listener's head toward the sound source and is expressed in the listener's head coordinate system.

\paragraph{Binaural feature extraction.}
We compute the short-time Fourier transform (STFT) of the left and right binaural signals using a $40\,\mathrm{ms}$ Hamming window, zero-padded to an FFT size of $2048$, and a hop size of $33.3\,\mathrm{ms}$, yielding a frame rate of $30\,\mathrm{Hz}$ aligned with the motion sequence. We keep the lowest $256$ frequency bins, corresponding to frequencies up to approximately $6\,\mathrm{kHz}$.

Let $L_{t,f}$ and $R_{t,f}$ denote the complex STFT coefficients of the left and right channels at time frame $t$ and frequency bin $f$. The SSL input contains four channels: (1) the average magnitude spectrum, (2-3) the sine and cosine of the interaural phase difference (IPD), and (4) the interaural level difference (ILD). The first channel encodes the overall spectral magnitude, the second and third channels encode the IPD without phase-wrapping discontinuities, and the fourth channel provides the relative loudness between the two ears. The resulting binaural feature is a tensor $\mathbf{X} \in \mathbb{R}^{4 \times 300 \times 256}$, where every sequence is represented by $300$ frames at $30\,\mathrm{Hz}$ to match the motion sequences. The extraction procedure is summarized in the Python pseudo-code below.

\begin{lstlisting}[style=pythonpseudo]
def extract_binaural_features(audio):           # audio: (2, N), 10 s @ 48 kHz

    # per-ear STFT: 40 ms Hamming window, hop of 33.3 ms (30 frames/s)
    S_L = librosa.stft(audio[0])                # (F, T), F = 1025 bins
    S_R = librosa.stft(audio[1])                # (F, T)

    # per-ear magnitude and phase
    A_L, phi_L = np.abs(S_L), np.angle(S_L)     # (F, T)
    A_R, phi_R = np.abs(S_R), np.angle(S_R)     # (F, T)

    # four binaural channels, each of shape (F, T)
    X1 = (A_L + A_R) / 2                        # (1) average magnitude
    X2 = np.sin(phi_L - phi_R)                  # (2) IPD sine
    X3 = np.cos(phi_L - phi_R)                  # (3) IPD cosine
    eps = 1e-10                                 # avoid division by zero
    X4 = 20 * np.log10((A_L + eps) / (A_R + eps)) # (4) ILD

    # stack the four channels, keep the lowest 256 bins, align with motion
    X = np.stack([X1, X2, X3, X4])              # (4, F, T)
    X = X[:, :256, :]                           # (4, 256, T)
    return X
\end{lstlisting}

\paragraph{Network architecture.}
The SSL network consists of a convolutional spectral encoder followed by a Transformer-based temporal encoder. The convolutional front-end contains three convolutional blocks. Each block applies a $3 \times 3$ convolution followed by a ReLU activation and max pooling along the frequency axis with stride $4$. This progressively reduces the frequency dimension from $256$ to $64$, $16$, and $4$, while preserving the temporal resolution.

After the convolutional blocks, the feature map is flattened at each time frame and projected into a $C$-dimensional latent space, where $C = 128$. We add a learnable absolute positional encoding to the frame-wise tokens before feeding them into the temporal encoder.

Different from~\citep{garcia2022binaural}, which adopts a GRU for temporal modeling, we use a Transformer encoder. The encoder has $L = 4$ layers, $8$ attention heads, and a feed-forward hidden dimension of $256$. We use pre-normalization Transformer blocks for stable training. The self-attention mechanism enables each frame to aggregate binaural evidence from the entire sequence, which helps resolve ambiguous localization cues caused by reverberation or transient source motion.

The Transformer outputs one token per frame. Each token is passed through two fully connected layers to regress a three-dimensional vector, which is then L2-normalized to obtain the predicted DoA:
$
\mathbf{d}_t
=
\frac{\mathbf{o}_t}{\|\mathbf{o}_t\|_2},
$
where $\mathbf{o}_t \in \mathbb{R}^{3}$ is the raw output vector before normalization.

\paragraph{Training objective.}
The SSL network is trained using head-relative source direction labels from \dbname and \dbnamessl. The ground-truth direction vector at frame $t$ is normalized to unit length and denoted as $\bar{\mathbf{d}}_t$. We optimize the cosine-distance loss:
$
\mathcal{L}_{\mathrm{ssl}}
=
\frac{1}{T}
\sum_{t=1}^{T}
\left(
1 - \mathbf{d}_t \cdot \bar{\mathbf{d}}_t
\right).
$
This objective is monotonically related to the angular error between the predicted and ground-truth DoA vectors and directly supervises the directional information used by the downstream motion generator.

During motion generation, the SSL network parameters are frozen. Given only binaural audio, the network produces the frame-wise DoA sequence
$
\mathbf{D} = \{\mathbf{d}_1, \ldots, \mathbf{d}_T\},
$
which is used as the spatial condition of the diffusion model.


\section{Evaluator Implementation Details} \label{appendix:evaluator}

\begin{figure}[t]
    \centering
    \includegraphics[width=\linewidth]{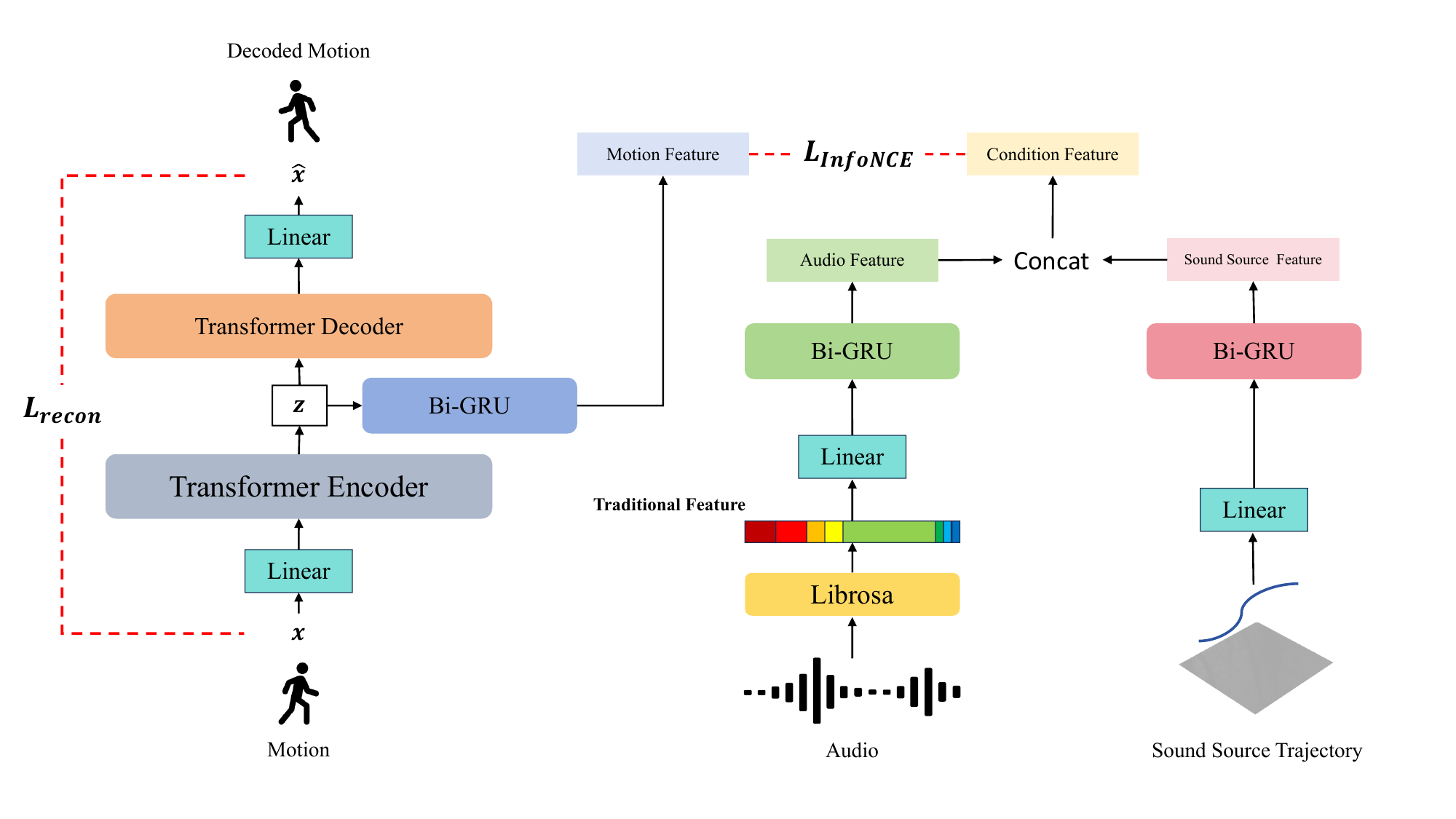}
    \caption{The evaluator framework consists of a motion autoencoder co-trained with Bi-GRU feature extractors. A reconstruction loss $\mathcal{L}_{\mathrm{recon}}$ is applied between the ground-truth motion $\mathbf{x}$ and the decoded motion $\hat{\mathbf{x}}$ produced by the motion autoencoder. Two Bi-GRU encoders extract the audio feature and the sound source feature from the binaural audio and the sound source trajectory, respectively, and the concatenation of the two features forms the condition latent. Additionally, an InfoNCE loss $\mathcal{L}_{\mathrm{InfoNCE}}$ operates on the extracted condition features and motion features.}
    \label{fig:evaluator}
\end{figure}

Inspired by the evaluator design from \citep{xu2025mospa}, our evaluator is a motion--condition matching network , which measures how well a motion sequence corresponds to a given spatial audio condition. It maps both modalities into a shared latent space, in which the R-precision metrics are computed by nearest-neighbor retrieval and the FID is computed between the feature distributions of the ground-truth and the generated motions. Different from~\citep{xu2025mospa}, whose condition only contains the acoustic content, our condition additionally involves the trajectory of the sound source, so that the evaluator is sensitive to the spatial correspondence between the moving sound source and the human reaction.

\paragraph{Network architecture.}
As illustrated in Fig.~\ref{fig:evaluator}, the evaluator consists of a motion autoencoder and three bidirectional GRUs (Bi-GRUs), each with 4 layers and a hidden size of 512. The feature of a Bi-GRU is obtained by concatenating its last forward and backward hidden states, which yields a 1024-dimensional feature vector.

The motion autoencoder follows a transformer encoder--decoder structure. The input motion sequence $\mathbf{x} \in \mathbb{R}^{T \times 272}$ is first projected to a 512-dimensional latent space and added with a positional encoding, and is then encoded by a 4-layer transformer encoder with 4 attention heads and a 1024-dimensional feed-forward layer. The encoder output is averaged over time into a single latent vector and replicated along the temporal axis, which serves as the query of a transformer decoder of the same configuration attending to the encoder output. A final linear layer maps the decoder output back to the 272-dimensional motion space to produce the reconstructed motion $\hat{\mathbf{x}}$. The output of the motion encoder is further encoded by a Bi-GRU to obtain the motion feature.

The condition branch consists of an audio encoder and a sound source encoder. The audio encoder takes the traditional features described in Appendix \ref{appendix:audio_features}. The sound source encoder takes the absolute position of the sound source in the world coordinate system $\mathbf{s} \in \mathbb{R}^{T \times 3}$, projects it to 64 dimensions and encodes it with a Bi-GRU. The resulting audio and sound source features are concatenated and projected by a linear layer into the condition feature, whose dimension matches that of the motion feature.

\paragraph{Training objective and details.}
The evaluator is optimized with two objectives. The motion autoencoder is supervised by a reconstruction loss $\mathcal{L}_{\mathrm{recon}}$, which encourages the latent space to preserve the motion content. The condition and motion representations are aligned through contrastive learning using the InfoNCE loss $\mathcal{L}_{\mathrm{InfoNCE}}$ \citep{oord2018representation}. During training, we randomly zero the sound source feature of an entire sequence with a probability of 0.3, producing audio-only conditions. This enables the evaluator to also apply to settings where the sound source trajectory is unavailable, while the audio condition is always retained.
We train the evaluator for up to 1{,}000 epochs using the Adam optimizer~\citep{kingma2014adam} with a learning rate of $5 \times 10^{-5}$ and a batch size of 32. The evaluator is trained on all sequences of \dbname.

\section{Evaluation Metrics} \label{appendix:metrics}

\textbf{Fréchet Inception Distance (FID).}
The FID is computed between the motion features of the ground-truth motions in the test set and those of the generated motions, where both sets of features are extracted by the frozen motion encoder of the evaluator. Gaussian distributions are fitted to the two feature sets, and the Fréchet distance between them serves as a metric for assessing the quality and fidelity of the synthesized motions.

\textbf{R-precision.}
We compute the R-precision within batches of 32 sequences of the test set, following the protocol of~\citep{guo2022generating}. For each batch, the conditions and the motion features are L2-normalized, and the cosine similarities between all condition--motion pairs are ranked, where the pool of each condition consists of its matched motion together with the other 31 in-batch motions as distractors. The R-precision is then obtained by determining the probability that the matched motion appears among the top-1, top-2 and top-3 ranked positions, and is averaged over all the batches. The metric is computed in both the condition-to-motion and the motion-to-condition retrieval directions. The query condition used in our experiments consists of the ground-truth binaural audio and the ground-truth sound source trajectory, independently of the condition adopted by each generator, so that all compared methods are evaluated against the same query set. The Real Motion row in the main tables corresponds to using the ground-truth motions as the retrieval target.

\textbf{Skating Ratio.}
We compute the skating ratio from recovered world-space foot joints, using the left and right foot joints indexed as 10 and 11 in the 22-joint SMPL protocol. For each foot, we compute its horizontal velocity on the ground plane at 30\,Hz and smooth it with a 5-frame window. A foot is considered \emph{planted} if it remains within 5\,cm above the floor in adjacent frames and its smoothed horizontal speed is below 0.30\,m/s. We count a valid contact frame as skating only when neither foot is planted. Airborne frames, where both feet are above 5\,cm, and penetration frames, where any foot is below the floor plane, are excluded from the skating ratio and reported separately as float and penetration ratios. The procedure is summarized as follows.

\vspace{4mm}
\begin{lstlisting}[style=pythonpseudo]
def skating_ratio(world_joints):                # (T, 22, 3) Z-up joints

    feet = world_joints[:, [10, 11], :]         # (T, 2, 3) left / right foot
    height = feet[:, :, 2]                      # (T, 2) foot height
    vel = calculate_feet_vel(feet)              # (T-1, 2) 
    vel = smooth(vel, window=5)                 # (T-1, 2)

    # a foot is planted when it stays near the floor and barely moves
    h = np.maximum(height[:-1], height[1:])     # (T-1, 2) max of two frames
    planted_feet = (h < 0.05) & (vel < 0.30)         # (T-1, 2)
    on_floor_feet = (h >= 0) & (h < 0.05)            # (T-1, 2) valid contact

    # a contact frame skates when neither foot is planted
    planted_frame = planted_feet.any(axis=1)           # (T-1,) any foot planted
    on_floor_frame = on_floor_feet.any(axis=1)         # (T-1,) any foot on floor
    skating_frame = ~planted_frame & on_floor_frame    # (T-1,) skating frame
    return skating_frame.sum() / on_floor_frame.sum()        # ratio over contact frames
\end{lstlisting}

Our skating ratio formulation is modified from GMD~\citep{karunratanakul2023gmd}, which marks a frame as skating when either foot is close to the ground and moves faster than $0.50$\,m/s. This can falsely flag frames where one foot is stably planted while the other foot briefly moves or drags near the floor, since the original metric evaluates each foot independently and ignores whether the body has a valid support foot. In contrast, our metric detects the absence of any stable support foot. This support-foot criterion avoids counting normal swing-foot motion as skating when the other foot stably supports the body.

To verify the metric behavior, we evaluate both definitions on the ground-truth motions of \dbname{}. As shown in Tab.~\ref{tab:skating_metric_gt}, the original formulation reports a skating ratio of 13.9\% on ground-truth data, which is inconsistent with the fact that these motions are captured from real humans and should contain little foot skating. In contrast, our formulation gives a much lower ratio of 1.26\%, indicating that the revised metric better reflects actual foot skating artifacts rather than false positives caused by normal foot motion or representation noise.

\begin{table}[h]
    \centering

    \begin{tabular}{l c}
    \toprule
    Metric Definition & Skating Ratio on GT $\downarrow$ \\
    \midrule
    GMD formulation~\citep{karunratanakul2023gmd} & 13.90\% \\
    Ours & 1.26\% \\
    \bottomrule
    \end{tabular}
    \caption{
    Skating ratio computed on the ground-truth motions of \dbname{} using different metric definitions.
    }
    \label{tab:skating_metric_gt}
\end{table}

\section{Traditional Audio Feature Set} \label{appendix:audio_features}

Following MOSPA \citep{xu2025mospa}, our acoustic feature set primarily includes \textit{Mel-frequency cepstral coefficients (MFCCs), MFCC delta, constant-Q chromagram, short-time Fourier transform (STFT) chromagram, onset strength, tempogram, and beats}~\citep{davis1980comparison, schorkhuber2010constant, grosche2010cyclic, mcfee2015librosa, bock2013maximum, ellis2007beat}. In addition, we incorporate the root mean square (RMS) energy $E_{rms}$ of the audio~\citep{mcfee2015librosa}, and define the active frames as $F_{active} = E_{rms} > 0.01$ to capture distance-related audio information.

The full list of audio features and their corresponding dimensions are listed in Tab.~\ref{tab:full_audio_features}. The audio is sampled at 48{,}000\,Hz. When computing frame-level audio features with Librosa, we set the FFT window length to 2048, the hop length to 1600, and use a Hann window, yielding audio features with a frame rate of 30\,Hz aligned with the motion sequences. The final shape of the audio features for a single ear is $\mathbb{R}^{T \times 335}$. We concatenate the features from the two ears along the feature dimension, obtaining the audio condition $\mathbf{a} \in \mathbb{R}^{T \times 670}$ for the evaluator, where $T=300$ denotes the number of frames in each motion sequence.

\begin{table}[h]
    \centering
    \setlength{\tabcolsep}{8pt}
    \renewcommand{\arraystretch}{1.2}
    \begin{tabular}{l | c}
        \hline
        \textbf{Feature Name}          & \textbf{Dimension} \\
        \hline
        MFCC                  & 20 \\
        MFCC delta            & 20 \\
        constant-Q chromagram & 12 \\
        STFT chromagram       & 12 \\
        onset strength        & 1 \\
        tempogram             & 267 \\
        one-hot beats         & 1 \\
        RMS energy            & 1 \\
        active frames         & 1 \\
        \hline
    \end{tabular}
    \vspace{2mm}
    \caption{The extracted audio features for a single ear comprise a $335$-dimensional vector. The total dimension of the binaural audio feature vector $\mathbf{a}$ is 670.}
    \label{tab:full_audio_features}
\end{table}

\section{Visualization of Reaction Genre Control}
\label{sec:appendix_genre_control}

\begin{figure}[h!]
    \centering
    \includegraphics[width=0.8\linewidth]{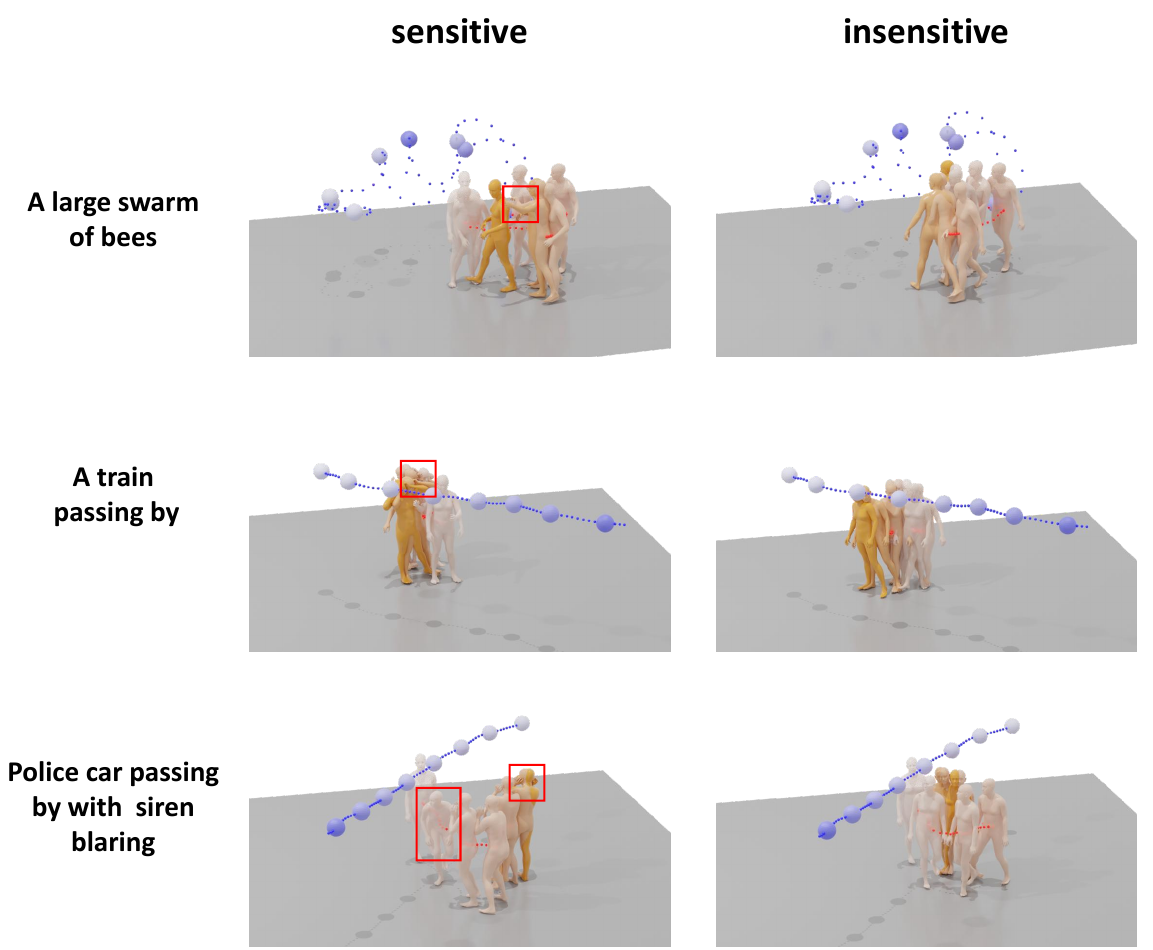}
    \caption{
    Qualitative comparison of reaction genre control on three audio examples. Given the same binaural audio, label 1, i.e., sensitive, produces stronger and larger-amplitude reactions than label 0, i.e., insensitive, demonstrating effective genre-based motion control.
    }

    \label{fig:genre_control}
\end{figure}

To evaluate the controllability of MOMO with respect to the reaction genre, we compare generated motions under two genre labels, where label 1 denotes a sensitive reaction and label 0 denotes an insensitive reaction. As shown in Fig.~\ref{fig:genre_control}, for three representative audio examples, we keep the binaural audio input unchanged and only switch the genre label during generation.

\name produces clearly different responses for the two labels. Motions generated with label 1 show larger amplitudes and stronger avoidance reactions, whereas those with label 0 are more restrained and subtle. These results demonstrate that MOMO can effectively control the reaction intensity and expressiveness through the genre condition.

\section{Robot Demonstrations}
\label{sec:appendix_robot_demo}

To investigate the applicability of dynamic spatial-audio-driven motion generation to embodied agents, we provide both simulated and real-world robot demonstrations. These demonstrations serve as preliminary evidence that the generated motions can be transferred to robotic platforms, suggesting the potential of our task formulation for future embodied deployment.

\begin{figure}[h]
    \centering
    \includegraphics[width=1.0\linewidth]{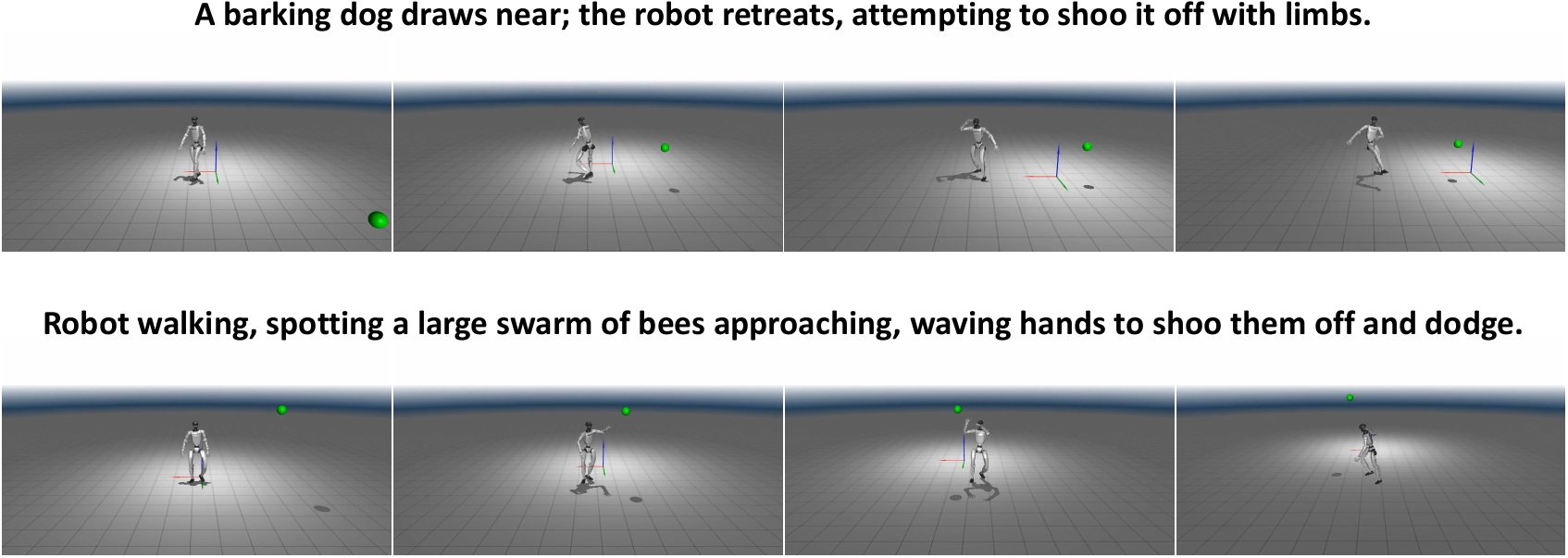}
    \caption{
    Simulated robot demonstration in MuJoCo. The generated SMPL-X motion is retargeted to a humanoid robot using GMR and tracked by a pretrained TWIST controller.
    }
    \label{fig:sim_robot_image}
\end{figure}

For the simulated robot demonstration, we retarget the generated SMPL-X motion parameters to a humanoid robot using GMR \citep{araujo2025gmr}. A pretrained TWIST controller \citep{ze2025twist} is then used to track the retargeted motion and drive the robot in the MuJoCo environment \citep{todorov2012mujoco}, as shown in Fig.~\ref{fig:sim_robot_image}.

For the real-world robot demonstration, we use SONIC \citep{luo2026sonic} as a general-purpose motion controller. The generated motion is fed into SONIC as the reference motion, and the real robot performs imitation based on this reference. The corresponding real-world demonstration is provided in the supplementary video. These results indicate that dynamic spatial audio-to-motion generation may serve as a useful interface for future robots to produce responsive behaviors toward moving sound sources. Please refer to our robot demonstration videos in the supplementary material.

\section{User Study} \label{appendix:user_study}

\begin{wrapfigure}{r}{0.5\textwidth}
    \centering
    \vspace{-4mm}
    \includegraphics[width=\linewidth]{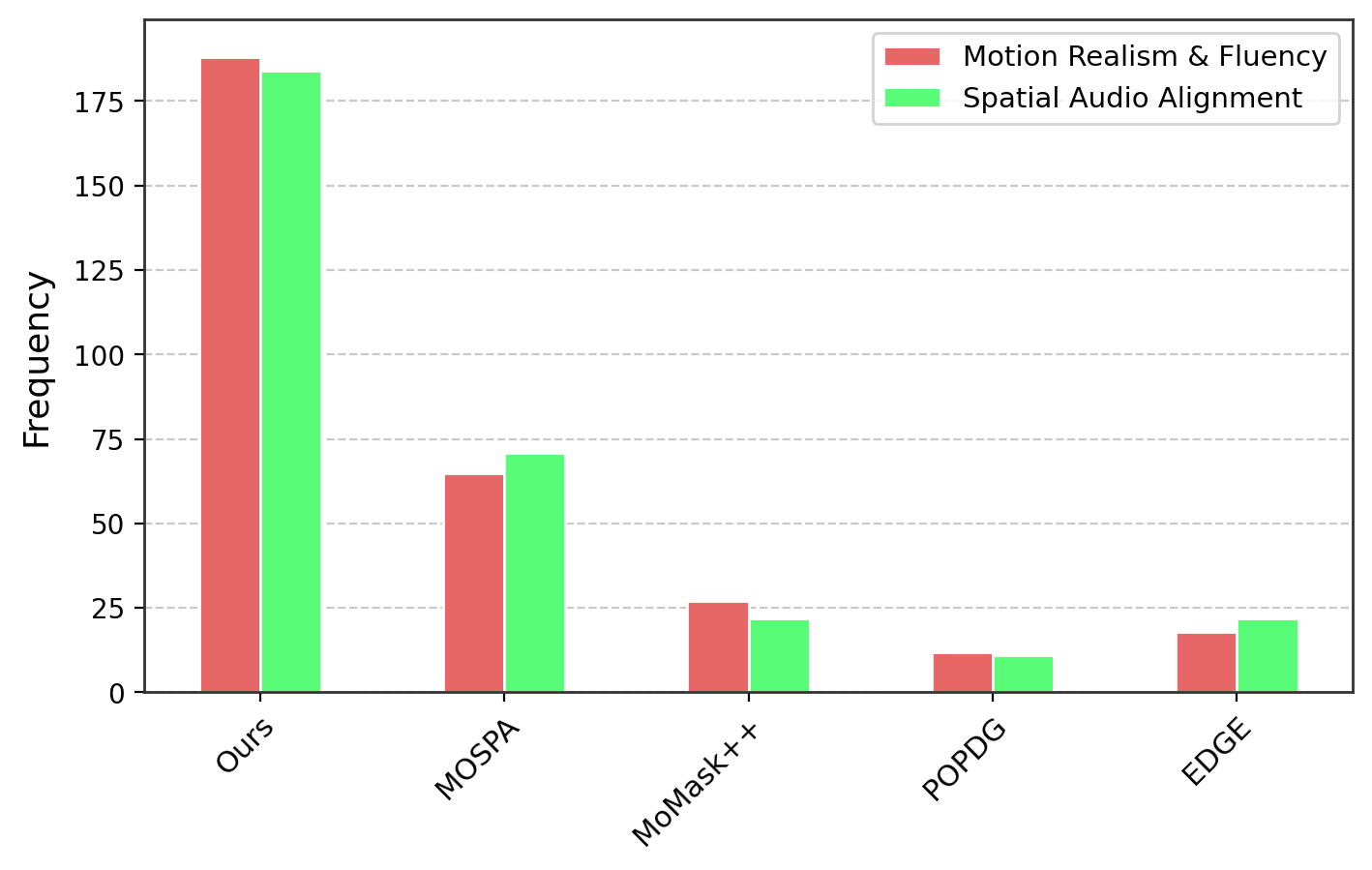}
    \vspace{-4mm}
    \caption{User study results.}
    \vspace{-4mm}
    \label{fig:user_study_best_frequency}
\end{wrapfigure}
We conduct a user study to evaluate the perceptual quality of generated motions. The study focuses on two aspects: 
(1) \textbf{motion realism and fluency}, which measures whether the generated motion looks natural, realistic, and temporally smooth; and 
(2) \textbf{alignment with dynamic spatial audio}, which measures whether the motion response is consistent with the audio content and the movement of the sound source.

\begin{figure}[h]
    \centering
    \includegraphics[width=0.95\linewidth]{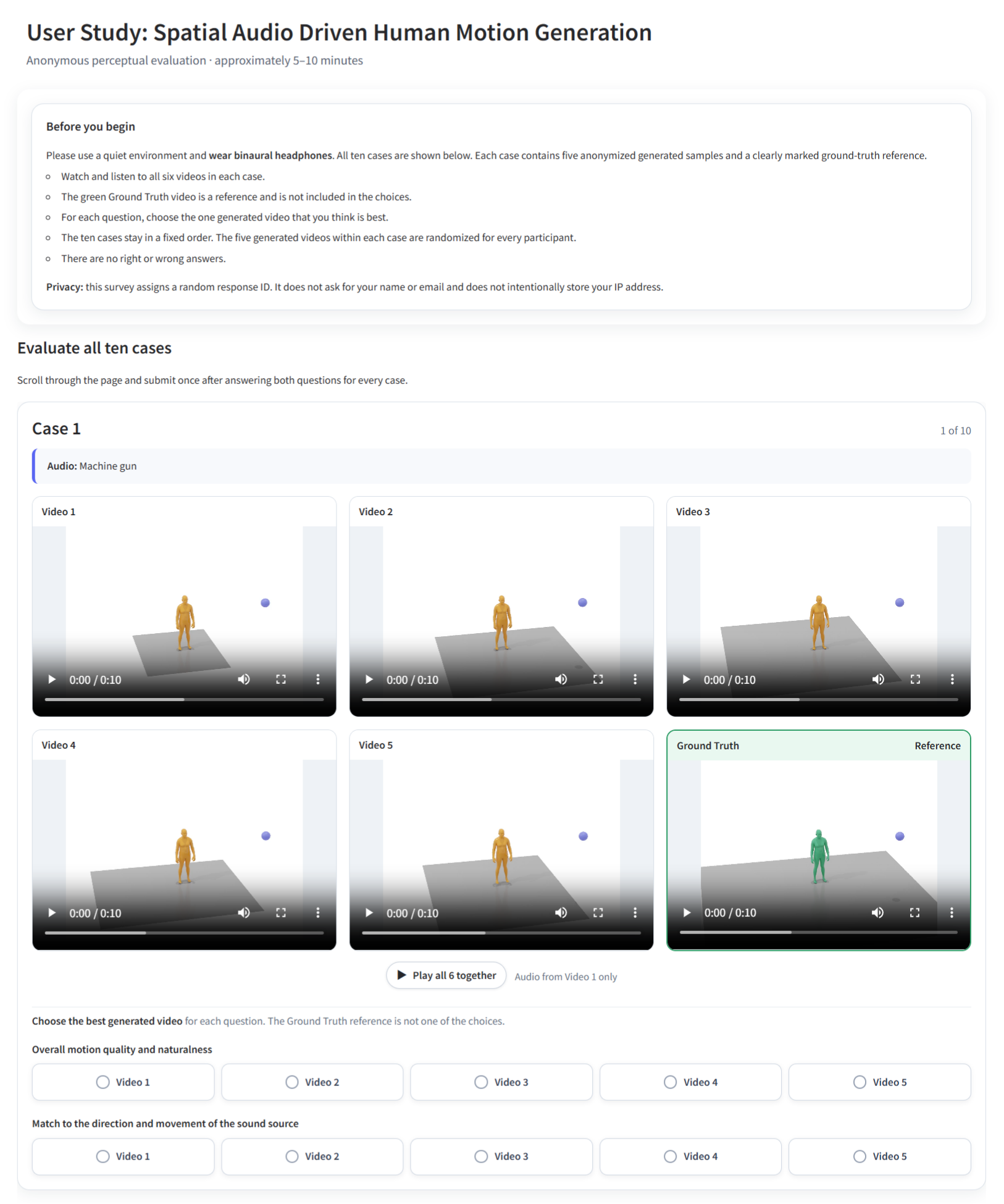}
    \caption{
    Screenshot of the user study interface.
    }
    \label{fig:user_study_interface}
\end{figure}

A total of 30 participants take part in the study. We provide 10 audio cases for evaluation. For each case, participants are shown six videos: one ground-truth motion video as the reference, and five generated motion videos from different methods, including \name{}(Ours), MOSPA~\citep{xu2025mospa}, MoMask++~\citep{hwang2025snapmogen}, POPDG~\citep{luo2024popdg}, and EDGE~\citep{tseng2023edge}. The order of the generated videos is randomized to reduce potential bias. The user study interface is shown in Fig.~\ref{fig:user_study_interface}.

For each audio case, participants watch the videos with the corresponding dynamic spatial audio and select the best generated result under each evaluation aspect. Fig.~\ref{fig:user_study_best_frequency} shows that \name{} is selected most frequently in both motion realism and fluency, and alignment with dynamic spatial audio, outperforming all baseline methods. This indicates that \name{} generates motions that are not only more natural and fluent, but also better aligned with dynamic spatial-audio cues.

